\documentclass[nohyperref]{article}
\usepackage[accepted]{icml2022}

\usepackage{microtype}
\usepackage{graphicx}
\usepackage{booktabs} 

\usepackage{url}            
\usepackage{amsfonts}       

\usepackage{tabularx}

\usepackage{hyperref}
\usepackage{amsmath, amsthm, amssymb, nicefrac, color}

\usepackage{dsfont} 

\usepackage{subcaption}

\usepackage[capitalize,noabbrev]{cleveref}
\usepackage[normalem]{ulem} 

\usepackage{thmtools}
\usepackage{thm-restate}
\declaretheorem[name=Theorem,numberwithin=section]{theorem}
\theoremstyle{definition}
\newtheorem{definition}{Definition}[section]
\theoremstyle{remark}
\newtheorem*{remark}{Remark}
\theoremstyle{plain}
\newtheorem{lemma}[theorem]{Lemma}
\newtheorem{claim}[theorem]{Claim}
\newtheorem{corollary}[theorem]{Corollary}
\newtheorem{fact}[theorem]{Fact}
\newtheorem{proposition}[theorem]{Proposition}

\newcommand{\R}{\mathbb{R}}

\newcommand{\calH}{\mathcal{H}}
\newcommand{\calX}{\mathcal{X}}
\newcommand{\calM}{\mathcal{M}}
\newcommand{\Loss}{\mathrm{L}}
\renewcommand{\S}{\mathsf{S}}
\newcommand{\T}{\mathsf{T}}

\newcommand{\pdfS}{\mathbb{P}_{\S}} 
\newcommand{\pdfT}{\mathbb{P}_{\T}} 
\newcommand{\pdim}{\mathrm{pdim}} 
\newcommand{\E}{\mathop{\mathbb{E}}}
\newcommand{\err}{\mathop{\mathrm{err}}}
\newcommand{\KL}[2]{\mathrm{KL}(#1\|#2)}
\newcommand{\TV}[2]{\mathrm{TV}(#1,#2)}
\newcommand{\chisq}[2]{\chi^2(#1\|#2)}
\newcommand{\Dinfty}[2]{\mathrm{D}^\infty(#1\|#2)}

\newcommand{\calP}{\Pr} 
\newcommand{\calN}{\mathcal{N}} 
\newcommand{\A}{\mathcal{A}} 

\newcommand{\1}{\mathds{1}} 
\newcommand{\ES}{\mathop{\mathbb{E}}\limits_{x\sim\S}} 
\newcommand{\ET}{\mathop{\mathbb{E}}\limits_{x\sim\T}} 

\renewcommand{\paragraph}[1]{\vspace{1mm}\noindent\textbf{#1}}

\icmltitlerunning{Transfer Learning In Differential Privacy's Hybrid-Model}

\begin{document}

\twocolumn[
\icmltitle{Transfer Learning In Differential Privacy's Hybrid-Model}



\icmlsetsymbol{equal}{*}

\begin{icmlauthorlist}
\icmlauthor{Refael Kohen}{biu}
\icmlauthor{Or Sheffet}{biu}
\end{icmlauthorlist}

\icmlaffiliation{biu}{Faculty of Engineering, Bar-Ilan University, Israel}

\icmlcorrespondingauthor{Refael Kohen}{refael.kohen@gmail.com}
\icmlcorrespondingauthor{Or Sheffet}{or.sheffet@biu.ac.il}

\icmlkeywords{Differential Privacy, Hybrid Model, Transfer Learning, Multiplicative Weights}

\vskip 0.3in
]


\printAffiliationsAndNotice{}  

\newcommand{\osnote}[1]{\textcolor{red}{#1}}

\begin{abstract}
The hybrid-model~\cite{AventKZHL17} in Differential Privacy is a an augmentation of the local-model where in addition to $N$ local-agents we are assisted by one special agent who is in fact a curator holding the sensitive details of $n$ additional individuals. Here we study the problem of machine learning in the hybrid-model where the $n$ individuals in the curator’s dataset are drawn from a \emph{different} distribution than the one of the general population (the local-agents). We give a general scheme~-- Subsample-Test-Reweigh~-- for this transfer learning problem, which reduces any curator-model DP-learner to a hybrid-model learner in this setting using iterative subsampling and reweighing of the $n$ examples held by the curator based on a smooth variation of the Multiplicative-Weights algorithm (introduced by~\citet{BunCS20}). Our scheme has a sample complexity which relies on the $\chi^2$-divergence between the two distributions. 
We give worst-case analysis bounds on the sample complexity required for our private reduction. Aiming to reduce said sample complexity, we give two specific instances our sample complexity can be drastically reduced (one instance is analyzed mathematically, while the other - empirically) and pose several directions for follow-up work.
\end{abstract}

\section{Introduction}
\label{sec:intro}

Differential privacy (DP) has become modern era's de-facto gold standard for privacy-preserving data analysis. In particular, the local model of DP~--- in which each user interacts in a computation by sending randomized messages whose view yields at most $\epsilon$-privacy loss~--- has gained much popularity due to its (relative) design simplicity. In particular, much work has dealt with the problem of machine learning in the local-model of DP: assuming the $N$ users' details are drawn i.i.d.~from some distribution $\T$, how can we design local-DP protocols for finding an hypothesis $h$ of small loss w.r.t.~$\T$. 
However, as opposed to the curator-model of DP~--- in which all data is held by a trusted curator in charge of executing the computation~--- machine learning in the local model of DP suffers from two drawbacks:  (1) its has a much larger sample complexity 
and (2) it is of limited learning capabilities~--- where only problems that are \textsf{SQ}-learnable are learnable in the local model, in contrast to the curator model that allows us to learn (almost) any \textsf{PAC}-learnable problem~\citep{KasiviswanathanLNRS08}.
In order to address these problems, there has been an extensive study in recent years regarding augmentations of DP's local-model. Most notably, the \emph{shuffle} model has gained much focus~\citep{BittauEMMRLRKTS17,CheuSUZZ19, BalleBGN19, BalleBGN20, GhaziG0PV21,Ghazi0MPS21} as it reduces the privacy-loss of local-model protocols drastically. Yet the focus of this work is on a different, far less studied,  augmentation of local-DP. 

We study the \emph{hybrid}-model of differential privacy~\cite{AventKZHL17}~--- in which a local-model computation protocol with $N$ users is augmented by the aid of one special agent who is in fact a curator holding the sensitive details of additional $n$ individuals. (We refer to the former $N$ users as the ``local-agents'' and the latter $n$ individuals as the ``curator-agents.'') The hybrid-model models a situation in which some users (the $n$ curator agents) trust a proposed curator and allow her unrestricted access to their sensitive details; while the remaining $N\gg n$ users trust solely themselves and opt for the local-model. 
Indeed, hybrid-model improves the learning capabilities of the local-model: the theoretical work of~\citet{BeimelKNSS20} have proven that in the hybrid-model certain problems, which are inefficiently solvable in the standard local-DP model, are efficiently and privately computable.
Alas, in the analysis of~\citet{BeimelKNSS20}, the $n$ curator-agents come from the \emph{exact same} distribution $\T$ as the remaining $N$ local-agents.\footnote{Such a situation may arise when an extrinsic powerful agency, e.g.\ the census, randomly samples $n$ individuals and mandates they provide the curator with their sensitive details.}
In contrast, this work is motivated by a setting in which the $n$ individuals \emph{voluntarily} opt-in to the curator-model. Coping with such a ``selection bias'' was posed by \citet{BeimelKNSS20} as an acute open problem since (quoting~\citet{BeimelKNSS20} verbatim:) ``from a practical point of view, this ... is aligned with current industry practices, and the ... individuals willing to contribute via a curator can be employees, technology enthusiasts, or individuals recruited as alpha- or beta-testers of products...~\cite{microsoft-opt-in, WindowsOptin, FirefoxOptin}''

In this work we model this selection bias as a particular type of a transfer learning problem. We no longer assume that the $n$ curator-agents were drawn from the \emph{same} distribution $\T$ as the local-agents, but rather that they were drawn from some \emph{different} distribution $\S$ (the \emph{source} distribution). Our goal, however, remains the same: to learn a good hypothesis w.r.t.~$\T$ (the \emph{target} distribution) while incurring at most $\epsilon$ privacy-loss for any of the agents involved in the computation. Naturally, should $\S$ and $\T$ be so different that they reside on disjoint support then the $n$ curator agents provide no assistance our transfer learning problem. Thus, in our model $\S$ and $\T$ are of bounded $\chi^2$-divergence (see details in Section~\ref{sec:preliminaries}). In other words, we study a particular variation of a \emph{transfer learning} problem: in the process of finding $h$ of small loss w.r.t.~$\T$ we are allowed to conduct a DP computation over a set of examples drawn from $\S$ and also to conduct a local-DP computation over $N$ additional examples drawn from $\T$. And so we ask:
\begin{center}
\unskip
\emph{Are there learning tasks that are infeasible in the local-model, yet can be computed privately and efficiently when we have a DP curator-model access to samples drawn from a \underline{different} distribution?}
\end{center}

\paragraph{Our Contribution.} Our transfer-learning problem has two na\"ive baselines. (1) Relying solely on the $N$ local-agents and learning a small loss hypothesis w.r.t.~$\T$ via some off-the-shelf local-model protocol; this is infeasible for certain problems (e.g. {\sf PARITY}~\cite{KasiviswanathanLNRS08}) and costly for others (e.g. sparse problem in a $d$-dimensional setting, where known sample complexity bounds are $\geq d$, \cite{DuchiJW13,SmithTU17}). (2) Relying solely on the $n$ curator-agents and learning an hypothesis via the some off-the-shelf DP learning algorithm guaranteed to return an hypothesis of small loss, such as a loss upper bounded by $\err_\S(h)\leq \nicefrac{\alpha} { \Dinfty{\T}{\S}}$ or $\err_\S(h)\leq \nicefrac{\alpha^2}{\chisq{\T}{\S}+1}$ (assuming these divergences are finite, see definition in Section~\ref{sec:preliminaries}).\footnote{This follows from the well known inequality, stating that for any event $E$ we have $\Pr_{\T}[E]\leq \sqrt{(\chisq{\T}{\S}+1)\Pr_{\S}[E]}$.} This venue is feasible only when indeed an hypothesis $h$ exists whose error over $\S$ is as small as we require; but fails to give a meaningful guarantee in an agnostic setting, even if the best hypothesis in $\calH$ has error $\Theta(\alpha)$. 

Our work proposes a third technique, whose sample complexity of the $N$ local-agents is independent of $d$ and whose sample complexity of the $n$ curator-agents depends on $pdim(\calH)$ (so if $|\calH|={\rm poly}(d)$, e.g. the Example in Section~\ref{sec:DP_version}, then $n={\rm polylog}(d)$) and which may be feasible in the agnostic setting as well.  Our proposed framework resembles `Subsample \& Aggregate'~\cite{nissim2007smooth} in broad brushstrokes, except that instead of subsampling in parallel in order to convert a non-private mechanism to a private one, we subsample sequentially in order to convert a curator-model learning mechanism over $\S$ to hybrid-model learner. The crux of our technique is that upon each time we subsample and produce an hypothesis of small-loss w.r.t.~$\S$ yet high loss w.r.t.~$\T$, we make a \emph{Multiplicative-Weights (MW)}-based update step to our subsampling distribution. With each MW-update we transition closer to a ``target'' distribution which is based on the \emph{importance sampling} (IS) weights of the points drawn from $\S$. We thus title our technique \emph{Subsample-Test-Reweigh}.


For clarity, we partition our analysis into two parts. The first, detailed in Section~\ref{sec:transfer_learning_non_private}, presents the main ideas of our techniques while tabling the notion of privacy for a later section. Specifically, seeing as we know that learning in the local-model is equivalent to \textsf{SQ}-learning and that the vast majority of \textsf{PAC}-learnable problems are learnt in the curator-model, we pose the following model. Fix an hypothesis $\ell$ which is \textsf{PAC}-learnable via hypothesis class $\calH$ of pdim $d$ through  some learning algorithm $\calM$. We are given $n$ labeled examples drawn from $\S$ and only a $\mathsf{SQ}$-oracle access to $\T$ (see formal definitions in Section~\ref{sec:preliminaries}). So we iteratively (1) set a distribution $\mu$ over the $n$ examples, (2) use $\calM$ to learn an hypothesis $h$ of small loss w.r.t.~$\mu$, (3) query the $\mathsf{SQ}$-oracle to see if $h$'s loss is sufficiently small (halt if yes), (4) if $h$ has large loss we reweigh the distribution using the MW-algorithm and proceed to the next iteration. In Section~\ref{sec:transfer_learning_non_private} we prove that w.h.p.~this algorithm outputs in $T=\tilde O(\alpha^{-2})$ iterations an hypothesis $h$ of error $\err_\T(h)=O(\alpha)$ provided $n=\tilde \Omega( d (\chisq{\T}{\S}+1)  \alpha^{-2} )$. The crux of the proof lies in showing (w.h.p.) the existence of a particular distribution $\bar u$ over the $n$ drawn samples from $\S$ s.t. $\forall h\in \calH$ the loss of $h$ w.r.t. $\bar u$ and its loss w.r.t.~$\T$ are close. Based on the seminal results of~\citet{CortesMM10}, it is not surprising that this $\bar u$ is the truncated IS weights $w(x)=\frac{\pdfT(x)}{\pdfS(x)}$ for each $x$ drawn from $\S$. 

Next, in Section~\ref{sec:DP_version} we give the privacy-preserving version of our MW-based technique from Section~\ref{sec:transfer_learning_non_private}. Replacing the \textsf{SQ}-oracle calls with simple applications of the Randomized-Response mechanism is trivial, but maintaining the privacy of the $n$ curator-model agents is far trickier. First, it is evident that our off-the-shelf learner $\calM$ must now be a privacy-preserving \textsf{PAC}-learner. More importantly, our MW-update step must also maintain bounded privacy-loss.  Luckily this latter point was already addressed  by~\citet{BunCS20} who use the notion of \emph{$\kappa$-dense distributions} to give a MW-based algorithm where any intermediate distribution $\mu$ is such that no one single point has probability mass exceeding $\nicefrac 1 {\kappa n}$ (see details in Section~\ref{sec:DP_version}). Using known several results regarding subsampling and privacy~\cite{KarwaV18, BunDRS18}, we end up with the following reduction. Denote $m_1$ as the sample complexity of some off-the-shelf curator-model learner that outputs an hypothesis of loss $\leq \alpha$ under privacy loss parameter of $\epsilon=1$. In order to successfully apply the learner $T$ times under our $\kappa$-dense MW-update subsampling scheme and have a total privacy-loss of $\epsilon$, we require a sample of $n$ curator-agents drawn from $\S$ where $n= \tilde \Omega(m_1\cdot  \frac{\sqrt T}{\epsilon\kappa})=\tilde \Omega(m_1\cdot  \frac{\chisq{\T}{\S}+1}{\epsilon\alpha^2})$ and $N = \tilde \Omega(\epsilon^{-2}\alpha^{-4})$ local-agents. While, admittedly, these bounds are less than ideal, this is the first result to prove the feasibility of private transfer learning using poly-size sample even for problems whose sample complexity in the local-model is exponential.

In Section~\ref{sec:conclusion} we pose suggestions as to how to reduce this bound as open problems, leveraging on the fact that the worst-case bounds for either the number of iterations $T$ {or}  the density parameter $\kappa$ may be drastically reduced  for specific hypothesis classes / instances. We give two particular examples of such instances: one proven rigorously (for the case of {\sf PARITY} under the uniform distribution) and one based on empirical evaluations of our (non private version of the) transfer-learning technique in a two high-dimensional Gaussian settings, in which our algorithm  makes far fewer iterations than our $O(\alpha^{-2})$ worst-case upper-bound.

\subsection{Related Work}
\label{subsec:intro_related_work}

The bounds and limitations of private learning were first established by~\citet{KasiviswanathanLNRS08} who proved that (roughly speaking) the learning capabilities of the local-model are equivalent to \textsf{SQ}-learning. This, together with the seminal result of~\citet{BlumFJKMR94}, gives an exponential lower bound on the number of local-agents required for learning {\sf PARITY} in the local-model (fully formally proven in~\citet{BeimelKNSS20}). Other results regarding the power and limitations of the local-model were given in~\citet{BeimelNO08,DuchiJW13,BassilyS15,SmithTU17,DuchiR19} culminating in~\citet{JosephMN019}. And yet, the classical SGD-algorithm is still applicable in the local-model~\cite{SmithTU17}. 

The two main works about the hybrid model~\cite{AventKZHL17,BeimelKNSS20} have been discussed already, as well as the elegant private boosting paradigm of~\citet{BunCS20}. Other works have also studied the applicability of the MW-algorithm for various tasks in DP~\cite{HardtR10,GuptaHRU11}. In the context of transfer learning, few works tied differential privacy with multi-task learning~\cite{GuptaRV16, XieBLZ17, LiKCT20}, hypthesis testing~\cite{WangGB18} and in a semi-supervised setting~\cite{Kumar22}; all focusing on empirical measuring of the utility and none giving any general framework with proven guarantees as this work. Also, it could be interesting to implement a version of our algorithm which isn't private w.r.t.~the samples from $\S$ in a ``PATE''-like setting~\cite{PapernotSMRTE18}
where public (or partially public) datasets are also available.

The classic problem of transfer learning has been studied extensively since the 20th century, and has too long and too rich of a history to be surveyed properly here.
We thus mention a few recent works that achieved sample complexity bounds based on importance sampling (IS) weights and show concentration bounds related to divergences between the source and that target distributions. Some works~\cite{agapiou2017importance,chatterjee2017sample} showed sample complexity bounds based, in part, on $\exp({\KL{\T}{\S}})\leq \chisq{\T}{\S}+1$ as well as other properties of the distribution\footnote{Namely, probability of drawing a point of large weight.} which lack our desired sub-Gaussian behavior. The seminal result of~\citet{CortesMM10} achieved a bound that is, in spirit, more similar to the desired sub-Gaussian behavior. They also proved that with sample size bounds depending on $\chisq{\T}{\S}$ we can achieve accurate estimation of the loss of any hypothesis (simultaneously) in a finite $\pdim$ hypothesis class ${\calH}$. Recently, \citet{metelli2021subgaussian} suggested a method of correction of the weights that allow obtaining sub-Gaussian behavior, assuming prior knowledge of the second moment of the weights. 
\citet{maia2022effective} studied the case where access to distribution $\T$ is restricted to unlabeled examples, and propose methods for evaluating the IS  weights. \citet{5539857} use a boosting algorithm (based on the MW algorithm) for transfer learning, yet in their work, the distributions $\S$ and $\T$ are identical but the features of the classes are different.

\section{Preliminaries}
\label{sec:preliminaries}

\paragraph{{\sf PAC}- and {\sf SQ}-Learning.} The seminal work of~\citet{Valiant84} defined {\sf PAC}-learnability, where the learner has access to poly-many examples drawn from a given distribution $\T$ over a domain $\calX$ labeled by some $\ell$ and its goal is to approximate $\ell$ as well as the best hypothesis in some given hypothesis class $\calH$. We measure our prediction success using some loss-function $\Loss$ where for every $x\in\calX, h\in \calH$ and any labelling function $\ell$ it holds that $\Loss(h(x),\ell(x))\in[0,1]$; and so for any distribution $\T$ we denote $\err_{\T}(h)=\E_{x\sim\T}[\Loss(h(x),\ell(x))]$. Thus, a $(\alpha,\beta)$-{\sf PAC}-learner outputs w.p.$\geq1-\beta$ an hypothesis $h\in{\cal H}$ where $\err_{\T}(h)\leq \alpha_{\calH}+\alpha$, where $\alpha_{\calH} \stackrel{\rm def}= \min_{h\in \calH}\{\err_\T(h)\}$. The learner's ability to draw $n$ i.i.d.~examples is simulated via an \emph{example-oracle} which upon a query returns such a labeled example. Thus, in the {\sf PAC}-mode, a learner has full access to all of the details of any drawn example. In contrast, the \emph{Statistical Query} ({\sf SQ}) model~\cite{Kearns98} restricts the operation of the algorithm to view solely statistical properties of the distribution $\T$ and the labeling function $\ell$. This is simulated via access to a \emph{${\sf SQ}_{\tau}$-oracle} which upon a statistical query $\phi:{\calX}\times\R \to[0,1]$ returns an estimation of $\E_{x\sim\T}[\phi(x,\ell(x))]$ up to a tolerance parameter $\tau$ (polynomially bounded away from $0$). 

\paragraph{Differential Privacy.} Given a domain $\calX$, two multi-sets $I,I'\in \calX^n$ are called \emph{neighbors} if they differ on a single entry. An algorithm (alternatively, mechanism) is said to be \emph{$(\epsilon,\delta)$-differentially private} (DP)~\cite{DworkMNS06,DworkKMMN06} if for any two neighboring $I,I'$ and any set $S$ of possible outputs we have: $\Pr[\calM(I)\in S]\leq e^\epsilon \Pr[\calM(I')\in S] + \delta$.

The Randomized-Response mechanism~\cite{Warner65,KasiviswanathanLNRS08} is one of the classic $(\epsilon,\delta)$-DP mechanisms in the local-model. Given privacy parameters $\epsilon,\delta$, on an input $b\in [0,1]$ it outputs $RR_{\epsilon,\delta}(b)\sim \calN(b,\frac{2\ln(2/\delta)}{\epsilon^2})$. When  applied to $N$ i.i.d.~draws from a Bernoulli r.v. of mean $\mu$, then we estimate $\mu$ by $\theta = \frac 1 N \sum_i RR_{\epsilon,\delta}(b_i)$.
Standard application of the Hoeffding bound and Gaussian concentration bounds proves that applying the mechanism to $N(\alpha,\beta,\epsilon,\delta)=\frac{4\ln(2/\delta)\ln(4/\beta)}{\epsilon^2 \alpha^2}$, we get that $\Pr[ |\theta-\mu| \leq\alpha]\geq 1-\beta$.

\emph{Differentially Private Machine Learning} is by now too large of a field to be surveyed properly. It was formally initiated by~\cite{KasiviswanathanLNRS08} who, as discussed above proved that the set of hypothesis-classes which is {\sf SQ}-learnable is conceptually equivalent to the set hypothesis-classes learnable in the local-model of DP. It is also worth mentioning the DP techniques for Empirical Risk Minimization and especially private SGD~\cite{ChaudhuriMS11,BassilyST14} which we use as our private learner in Appendix~\ref{sec:privacy_sgd}.

\paragraph{Divergence Between Distributions.} Given two distributions $\S$ and $\T$ over the same domain $\calX$ that have a Radon-Nikodym derivative, we denote said derivative as the \emph{importance sampling (IS) weight} at a point $x\in {\calX}$ as $w(x)=\frac{\pdfT(x)}{\pdfS(x)}$.
Given a convex function $f:(0,\infty)\to \R$ where $f(1)=0$, the $f$-divergence between two distribution $\T$ and $\S$ is $D^f(\T||\S) = \E_{x\sim \S}[f(w(x))]$. In the specific case when $f(x) = \tfrac 1 2|x-1|$ we obtain the \emph{total variation} distance, denoted $\TV{\T}{\S}$; when $f(x)=x\log(x)$ we obtain the \emph{Kullback-Leibler} (KL) divergence, denoted $\KL{\T}{\S}$; and when $f(x)=x^2-1$ we obtain the \emph{$\chi^2$-divergence}, denoted $\chisq{\T}{\S}$. Thus, a finite $\chi^2$-divergence between distributions implies a finite second moment for $w(x)$. It is indeed quite simple to see that $\E_{x\sim \T}[w(x)]=\E_{x\sim \S}[w(x)^2]=\chisq{\T}{\S}+1$. Moreover, a well-known result~\cite{CsiszarS04} states the for any twice-differentiable $f$ it holds that $D^f(\T\|\S)\approx \frac{f''(1)}{2}\chisq{\T}{\S}$ (as follows from $f$'s Taylor series). In the context of transfer learning,~\citet{CortesMM10} proved that weighing all examples in a sufficiently large sample drawn from $\S$ according to their IS weights gives an accurate estimation of the loss of any hypothesis in a finite pdim hypothesis class ${\calH}$ over $\T$; where their sample size bounds depend on $\chisq{\T}{\S}$.
Another well-used notion of divergence is the $\alpha$-Reyni divergence between $\T$ and $\S$, definted as $D_\alpha(\T\|\S)=\frac 1 {\alpha-1}\ln(\E_{x\sim\S}[w(x)^\alpha])$ for any $\alpha>1$, which was also used to define similar notions of privacy~\cite{BunS16,Mironov17,BunDRS18}. Note that $D_2(\T\|\S) = \ln(\chisq{\T}{\S}+1)$. We also denote $\Dinfty{\T}{\S}=\sup_{x\in {\calX}}w(x)$.
 
\paragraph{Bernstein Inequality:} In our work we {use} several standard concentration bounds (Markov- / Chernoff- / Hoeffding-inequalities), and also the slightly less familiar inequality of \citet{Bernstein54}:
Let $\{X_i\}_{i=1}^n$  be independent zero-mean random variables. Suppose that $|X_i| \le M$ almost surely, for all $i$. Then for any positive $t$,
\begin{equation}
      \calP \bigg(\sum_i X_i \ge t  \bigg) \le \exp \bigg(\frac{-t^2}{2\sum_i \E[X_i^2] + 2tM/3}\bigg)  
      \label{eq:bernstein-ieq}
\end{equation}

\section{A Non-Private Model}
\label{sec:transfer_learning_non_private}

For the sake of clarity, we introduce our algorithm in stages~--- where first we disregard the privacy aspect of the problem. In this section we deal with a specific transfer learning model, whose details are as follows. We are given a known domain $\calX$ and some unknown labelling function $\ell:\calX\to \R$. We also have the following access oracles to two unknown distributions $\S$ and $\T$ over $\calX$~--- we are given an \emph{example oracle} $\mathsf{Ex}$ access to $\S$, that upon a query returns an example $x$ drawn from $\S$ and labeled by $\ell$; and we are given a \emph{statistical query oracle} $\mathsf{SQ}_\tau$ to $\T$, that upon a query $\phi:\calX\times \R\to [0,1]$ returns an answer in the range $\E_{x\sim\T}[\phi(x,\ell(x)]\pm\tau$. 
Our goal is learn $\ell$ through some hypothesis class $\calH$; i.e. to find an hypothesis $h$ whose loss w.r.t.$~\ell$ over $\T$ is comparable to the loss of the best hypothesis in $\calH$, whose ${\pdim}(\calH)=d$. Formally, we introduce our algorithm in the realizable setting, so given a parameter $\alpha>0$ and a loss function $\Loss:\R^2\to [0,1]$ our goal is to find $h$ such that $\err_\T(h) = \alpha$. (We later discuss extension to the agnostic case.)
In addition, we are given an algorithm $\calM$ which is a $(\alpha,\beta)$-{\sf PAC} learner for our hypothesis class $\calH$. Namely, given $\alpha,\beta>0$, our learner $\calM$ takes a sample of $m(\alpha,\beta)$ i.i.d.~examples drawn from some distribution $\mu$ and labeled by some $\ell$ and w.p. $\geq 1-\beta$ returns a function $h\in \calH$ whose loss is upper bounded by $\err_\mu(h) \leq \alpha$. So, had we been able to draw i.i.d.~examples from $\T$, we would have just fed them to the algorithm and produce a good hypothesis $h$. Alas, in our model, we only have statistical query oracle  access to $\T$, $\mathsf{SQ}_\tau$, with error of $\pm\tau$.


In this section, we propose a general reduction of a {\sf PAC}-learner over $\S$ to finding a good hypothesis w.r.t.~$\T$, which depends solely on $\chisq{\T}{\S}\stackrel{\rm def}{=}\chi^2$.
We argue that  Algorithm~\ref{alg:non_private_transfer_learner} achieves our goal within $T=O(\log(\nicefrac {\chi^2+1} \alpha)/\alpha^2)$-iterations. 

Note that Algorithm~\ref{alg:non_private_transfer_learner} is presented for the realizable case. If we deal with an agnostic case, namely~--- where  $\alpha_\calH>0$, then our algorithm iterates as long as $\err_{\mu^t}(h^t) < \alpha_{\calH}+\alpha$. (A condition which ought to hold when we begin with $h^1$, the a good hypothesis w.r.t.~$\S$ and $\err_{\S}(h^1) \ll \err_{\T}(h^1)$.) It ought to be clear that by returning $h^t$ with the smallest error estimation given by the $\mathsf{SQ}$-oracle in all $T$ iterations, an hypothesis $h\in \calH$ of small loss w.r.t.~$\T$ is obtained. 

\begin{algorithm}[tb]
   \caption{Non-Private Subsample-Test-Reweigh}
   \label{alg:non_private_transfer_learner}
\begin{algorithmic}
   \STATE {\bfseries Input:} parameters $0<\alpha,\beta<\nicefrac 1 8$
   \STATE Draw $n\geq \frac{800(\chi^2+1)\log(\nicefrac 1 \alpha)}{\alpha^2}(d\log(\tfrac {400d} {\alpha^3})+\log(\tfrac 8 \beta))$ labeled points i.i.d.~from $\S$, denoted $x_1, x_2, ..., x_n$.
   \STATE Set weight $w_i^1\gets 1$ for each $1\leq i \leq n$.
   \STATE Set $T\gets\frac{32\log_2(\frac{8(\chi^2+1)}{\alpha})}{\alpha^2}$.
   \FOR {($t = 1,2,3,...,T$)}
   \STATE Set $\mu^t$ as a distribution where $\mu_i^t \propto w_i^{t}$.
   \STATE Draw $m(\alpha, \nicefrac \beta T)$ examples i.i.d~from $\mu^t$.
    \STATE Apply $\calM$ to the drawn points and obtain a function $h^t$.
    \STATE Set $a^t$ as the reply of the $\mathsf{SQ_\tau}$-oracle for the query $\phi^t(x,\ell(x))~=~\Loss(h^t(x),\ell(x))$. 
   \IF {($a^t > 2\alpha +\tau+\alpha_\calH$)}
   \STATE $\forall i$ set $w^{t+1}_i \gets w_i^t \cdot \exp(-\nicefrac \alpha 8 \cdot [1-\Loss(h^t(x_i),\ell(x_i))])$
   \ELSE
   \STATE {\bfseries return} $h^t$ (and halt)
   \ENDIF
   \ENDFOR
\end{algorithmic}
\end{algorithm}

\begin{theorem}
\label{thm:non_private_MW_correct}
In the above-described setting, w.p.$\geq 1-2\beta$ Algorithm~\ref{alg:non_private_transfer_learner},  halts and outputs an hypothesis $h$ with $\err_\T(h)\leq  2\alpha+2\tau$.
\end{theorem}
The proof of Theorem~\ref{thm:non_private_MW_correct} relies on proving the following lemma, and builds on~--- and extends~--- a theorem of~\citet{CortesMM10}.
\begin{lemma}
\label{lem:exist_capped_weights}
Given $0<\alpha,\beta\leq \nicefrac 1 8$ and two distributions $\S$ and $\T$ whose $\chi^2$-divergence is $\chi^2(\T||\S)=\chi^2$. Then, if $n=\Omega((\chi^2+1)\frac{\log(\nicefrac 1 \alpha)}{\alpha^2}(d\log(\tfrac {d} {\alpha})+\log(\tfrac 1 \beta)))$ w.p. $\geq 1-\beta$ it holds that there exists a distribution $\bar u$ over the $n$ drawn points such that (i) its divergence to the uniform distribution over the $n$ drawn point, $U_{[n]}$, satisfies $\KL{\bar u}{U_{[n]}}\leq \log_2(\nicefrac{8(\chi^2+1)}{\alpha})$ and also $\textrm{(ii) } \forall h\in \calH, ~ \err\nolimits_u(h)\geq \err\nolimits_\T(h) -\nicefrac {5\alpha} 8$.
\end{lemma}

\begin{proof}[Proof of Theorem~\ref{thm:non_private_MW_correct}]
Based on Lemma~\ref{lem:exist_capped_weights}, we now simply apply the characterization of the MW-algorithm from the seminal work of~\citet{AroraHK12}. Setting the `cost' of example $i$ w.r.t.~hypothesis $h^t$ as $m^t_i = 1-\Loss(h^t(x_i),\ell(x_i))$ we have that applying the MW-algorithm with costs $\bar m^1, ..., \bar m^T$ and with an update rate of $\eta=\nicefrac\alpha 8$, we get that for any fixed distribution $\nu$ it holds that
\[ \sum_t \E_{i\sim\mu^t}[m_i^t] \leq \sum_t \E_{i \sim \nu}[m_i^t] + \frac {\alpha T} 8 + \frac {8\KL{\nu}{U_{[n]}}}{\alpha}  \]
Note that w.p. $\geq 1-\beta$ our algorithm $\calM$ returns an hypothesis of $\err_{\mu^t}\leq \alpha$ in all $T$ iterations. In contrast, we know that each MW-update happens since $\err_\T(h^t)\geq 2\alpha+\tau-\tau = 2\alpha$; and so, w.p. $\geq 1-\beta$, we have a distribution $\nu=\bar u$ given by Lemma~\ref{lem:exist_capped_weights}, which we plug-in to the above equation and have
\[\resizebox{0.49\textwidth}{!}{$ \sum\limits_t (1-\alpha)\leq \sum\limits_t [1-(2\alpha-\frac {5\alpha} 8)] +   \frac {\alpha T} 8 + \frac{8\log_2(\frac{8(\chi^2+1)}{\alpha})}{\alpha} $} \]
Rearranging we get the bound
$\frac{3\alpha T }8 \leq \frac {\alpha T} 8 + \frac{8\log_2(\frac{8(\chi^2+1)}{\alpha})}{\alpha}$.
Hence, w.p.$\geq 1-2\beta$ Algorithm~\ref{alg:non_private_transfer_learner} halts within some iteration $t^*\leq \frac{32\log_2(\frac{8(\chi^2+1)}{\alpha})}{\alpha^2}$ steps, which means $\err_\T(h^{t^*}) \leq (2\alpha+\tau)+\tau = 2\alpha+2\tau$.
\end{proof}

\begin{proof}
[Proof of Lemma~\ref{lem:exist_capped_weights}] 
Let $w(x):\calX\to \R_+$ be the function defined by $w(x) = \frac{\pdfT(x)}{\pdfS(x)}$, where $\pdfT$ and $\pdfS$ denotes the PDFs of $\T$ and $\S$ resp. We call $w(x)$ the \emph{weight of $x$}. Lets $u(x) = \min (w(x), \frac{4(\chi^2+1)}{\alpha})$ be the \emph{truncated weight} of $x$.
Recall that $\E\limits_{x\sim\S}[w(x)]=1$ and that $\E\limits_{x\sim \S}[w(x)^2]=\E\limits_{x\sim \T}[w(x)]=\chi^2+1$. Thus, since for every $x\in\calX$ it holds that $0\leq u(x)\leq w(x)$ then we have $\E\limits_{x\sim\S}[u(x)]\leq 1$ and that $\E\limits_{x\sim \S}[u(x)^2]  \leq \E\limits_{x\sim \T}[u(x)]\leq \chi^2+1$.  Denoting $\A = \{x\in \calX:~ w(x)> \frac{4(\chi^2+1)}{\alpha}\}$, we can apply the Markov inequality to infer that $\Pr\limits_{x\sim \T}[x\in \A] = \E\limits_{x\sim \T}[\1_\A(x)] \leq \nicefrac \alpha 4$ with $\1_\A(x)$ denoting the indicator of whether $x\in \A$ or not. And so:
\begin{align*}
    &\E_{x\sim \S}[w(x)-u(x)] = \E_{x\sim \S}[(w(x)-u(x))\1_\A(x)]
    \cr &\qquad\leq \int_\calX w(x)\1_\A(x) \pdfS(x)dx 
    \cr &\qquad= \int_{\calX}\1_\A(x)\pdfT(x)dx = \E_{x\sim T}[\1_\A(x)] \leq \nicefrac \alpha 4
\end{align*}
which implies that $\E_{x\sim \S}[u(x)]\geq 1-\nicefrac \alpha 4$. 

We continue to bounding $U = \sum_{i=1}^n u(x_i)$ for the $n$ points $x_1, x_2,..., x_n$ in our sample taken i.i.d.~from $\S$, provided $n\geq \frac{150(\chi^2+1)\ln(\nicefrac 4\beta)}{\alpha^2}$. To that end, we apply the Bernestein inequality. First, we have that
\begin{align*}
    &\Pr[ U > (1 + \nicefrac{\alpha}8)n ] \leq  \Pr[ U-n\E[u(x_i)]> \nicefrac{\alpha n}8]
    \cr&~~~~ \leq \exp\left( - \frac{\nicefrac{\alpha^2n^2}{8^2}}{ 2\sum_i \textrm{Var}[u(x_i)] + \frac 2 3 (\frac{4(\chi^2+1)}{\alpha})\cdot  \nicefrac{\alpha n}8 } \right)
    \cr &~~~~ \leq \exp\left( - \frac{\alpha^2n^2}{ 128n (\chi^2+1) + \frac {64n} 3 (\chi^2+1) }\right)
    \cr &~~~~ \leq \exp\left( - \frac{\alpha^2n}{ 150 (\chi^2+1) }\right) \leq \nicefrac \beta 4
\end{align*}
and similarly we can prove $\Pr[ n(1-\nicefrac \alpha 4) - U > \nicefrac{\alpha n}8] \leq\nicefrac \beta 4$. 
Hence, w.p. $\geq 1-\nicefrac \beta 2$ it holds that $(1 - \nicefrac{3\alpha}8)n\leq U\leq (1 + \nicefrac{\alpha}8)n$.

Now, setting $\bar u$ as the distribution where point $i$ is sampled w.p. $\frac{u(x_i)}{U}$ we can use the above bound on $U$ to infer that
\begin{align*}
    \KL{\bar u}{U_{[n]}} &= \sum_i \tfrac{u(x_i)}{U} \log_2( {\frac{u(x_i)}{U}}/{\frac 1 n})
    \cr &\leq  \sum_i \tfrac{u(x_i)}{U} \log_2\left(\frac{u(x_i)} {1-\nicefrac{3\alpha}8}\right) 
    \cr &\leq \sum_i \tfrac{u(x_i)}{U}  \log_2(\tfrac{4(\chi^2+1)}{\alpha} \cdot \tfrac {8} {5}) 
    {\leq} \log_2(\tfrac{8(\chi^2+1)}{\alpha})
\end{align*}
{proving the first part of the claim.}
As for the second part of the claim, we simply use the result of~\citet{CortesMM10} which shows universal convergence w.r.t.~any unnormalized weights function. Namely, by setting $\alpha' = \frac{\alpha}{\sqrt{50(\chi^2+1)\log(1/\alpha)}}$,\footnote{Which satisfies that $\nicefrac \alpha 4 > \alpha'\sqrt{(2+\ln(\nicefrac{1}{\alpha'}))(\chi^2+1)} = \alpha'\sqrt{(2+\ln(\nicefrac{1}{\alpha'}))\E_{x\sim S}[w^2(x)]}$, when $\alpha \leq \nicefrac 1 8$.} we have that for any hypothesis class $\calH$ of $\textrm{pdim}(\calH)=d$, the following holds
\begin{align*}
    &\resizebox{\textwidth/2}{!}{$\Pr[\sup\limits_{h\in\calH} \big\{\E\limits_{x\sim \S}[u(x)\Loss(h(x),\ell(x))]- \sum_i \frac{u(x_i)}n\Loss(h(x_i),\ell(x_i))\big\} > \tfrac \alpha 4  ]$} \cr
    &\resizebox{\textwidth/2}{!}{$ \leq \Pr[\sup\limits_{h\in\calH} \big\{\frac{\E\limits_{x\sim \S}[u(x)\Loss(h(x),\ell(x))]- \sum_i \frac{u(x_i)}n\Loss(h(x_i),\ell(x_i))}{\sqrt{\E_{x\sim \S}[w^2(x)\Loss^2(h(x),\ell(x))]}}\big\} > \alpha'\sqrt{2+\log(\tfrac 1 {\alpha'})}  ]$} \cr
    & \resizebox{\textwidth/2}{!}{$\leq \Pr[\sup\limits_{h\in\calH} \big\{\frac{\E\limits_{x\sim \S}[u(x)\Loss(h(x),\ell(x))]- \sum_i \frac {u(x_i)}n\Loss(h(x_i),\ell(x_i))}{\sqrt{\E_{x\sim \S}[u^2(x)\Loss^2(h(x),\ell(x))]}}\big\} > \alpha'\sqrt{2+\log(\tfrac 1 {\alpha'})}  ] $}
    \cr &\stackrel{(\ast)}\leq 4\exp\left( d\log(\tfrac{2en}{d}) -\tfrac{n\alpha'^2}{4} \right)
    \cr &= 4\exp\left( d\log(\tfrac{2en}{d}) -\tfrac{n\alpha^2}{200(\chi^2+1)\log(1/\alpha)} \right) \leq \tfrac\beta 2
\end{align*}
when $n\geq \frac{800(\chi^2+1)\log(\nicefrac 1 \alpha)}{\alpha^2}(d\log(\tfrac {400d} {\alpha^3})+\log(\tfrac 8 \beta)) $. Note that $(\ast)$ is taken verbatim from~\citet{CortesMM10} Theorem 8. Thus, w.p. $\geq 1-\beta$ both the above bounds on $U$ hold and we have that for any $h\in \calH$ it holds that
\begin{align*}
    \err_\T(h) &= \E_{x\sim \T}[\Loss(h(x),\ell(x))] = \E_{x\sim \S}[w(x)\Loss(h(x),\ell(x))]
    \cr & \leq \E_{x\sim \S}[u(x)\Loss(h(x),\ell(x))] + \E_{x\sim \S}[w(x)-u(x)] 
    \cr &\leq  \Big[\tfrac \alpha 4 +\tfrac 1 n \sum_i u(x_i)\Loss(h(x_i),\ell(x_i))\Big] + \tfrac \alpha 4
    \cr & \leq \tfrac {\alpha}2 + \tfrac {(1+\nicefrac \alpha 8)} U \sum_i u(x_i)\Loss(h(x_i),\ell(x_i))
    \cr &\leq \tfrac{5\alpha} 8 + \E_{x_i\sim \bar u}[\Loss(h(x_i),\ell(x_i)]\qedhere
\end{align*}
\end{proof}

\section{The Private Boosting Paradigm}
\label{sec:DP_version}

We now turn our attention to the full hybrid-model, and to our need to privatize Algorithm~\ref{alg:non_private_transfer_learner}. In order to design a private version of Algorithm~\ref{alg:non_private_transfer_learner} one required multiple changes. First, and perhaps the easiest, is the fact that instead of a $\mathsf{SQ}_\tau$ oracle access to $\T$ we estimate the error of $h^t$ using RR (in the local-model). Assuming our algorithm makes at most $T$ iterations, standard argument shows that each such query requires $\Omega(\frac 1 {\epsilon^2\alpha^2} \log(T/\beta) )$ users so that w.p. $\geq 1-\nicefrac\beta T$ we get a $\alpha$-esimation of $\err_T(h^t)$; so, the number of local-users required for our paradigm is $\Omega(\frac {T\log(T/\beta)} {\epsilon^2\alpha^2}) = \Omega(\frac{\log((\chi^2+1)/\alpha)\log(\log(\chi^2+1)/\alpha\beta)}{\epsilon^2\alpha^4})$.

Second, which is also a rather straight-forward change, is that we need to replace the learning mechanism $\calM$ with a \emph{privacy-preserving} learning mechanism whose sample complexity depends also on the privacy-loss parameter(s). This implies that the sample complexity of $\calM$ is a function of the 4 parameter $m=m(\alpha,\beta,\epsilon,\delta)$.\footnote{For brevity, we use $(\epsilon,\delta)$ as the privacy parameters, even if $\calM$ is a (possibly truncated) zCDP-mechanism or R\'eyni-DP.} The question of setting the privacy parameters of $\calM$, thereby inferring the sample complexity of $\calM$, will be discussed momentarily. 

Lastly, the more challenging aspect of the problem is maintaining the privacy of the samples among the $n$ examples that are drawn from $\S$. To that end, we rely on the MW-variant of~\cite{BunCS20}, which in turn requires we introduce one more parameter, namely $\kappa$, into the problem.

\begin{definition}
\label{def:kappa_dense_distribution}
Fix some $0<\kappa<1$. Given $n$ points and a set of weights $w_1, w_2, ..., w_n \geq 0$, we denote $w_{\rm avg} = \tfrac 1 n \sum_i w_i$ and $w_{\max}=\max_i \{w_i\}$. We say that the distribution induced by these weights, namely the distribution where $\mu_i \propto w_i$, is \emph{$\kappa$-dense} if $\kappa w_{\max}\leq  w_{\rm avg}$.
\end{definition}

From a privacy stand-point, it is clear why a dense distribution is a desired trait: it makes it so that in a random sample of $m$ draws from $\mu$ we expect each point $i$ to be drawn no more than $1/\kappa$ times. \cite{BunCS20} showed that for any set of weights $\bar w = (w_1, w_2, ..., w_n)$ where $\forall i,~ w_i \in (0,\kappa]$, by setting
\begin{equation}
    \Pi_\kappa(\bar w) = \resizebox{0.35\textwidth}{!}{$\big(\min\{  c\cdot w_1, 1\}, \min\{  c\cdot w_2, 1\}, \ldots, \min\{  c\cdot w_n, 1\}\big)$}    \label{eq:projection_kappa_dense}
\end{equation}
for the smallest $c$ s.t. $\|\Pi_\kappa(w)\|_1=\kappa n$, we obtain a set of weights whose induced distribution $\mu = \frac{\Pi_\kappa(\bar w)}{\|\Pi_\kappa(\bar w)\|_1}$ is $\kappa$-dense. 
Using this projection we get the following claim.

\begin{claim}
\label{clm:sampling_into_DP}
Let $S$ and $S'$ be any two neighboring datasets of size $n$ each. Let $\bar w$ and $\bar w'$ be two weight vectors in $(0,\kappa]^n$ which may differ on the only entry that differs between $S$ and $S'$, and let $\mu$ and $\mu'$ be the two distributions derived from $\bar w$ and $\bar w'$ resp.~using the projection of~\eqref{eq:projection_kappa_dense}. Let $\calH$ be an hypothesis class and let $\calM$ be a $(\epsilon,\delta)$-DP mechanism that takes as input a dataset of size $m=m(\alpha,\beta,\epsilon,\delta)$ and outputs some $h\in \calH$. Then, for any $T\subset \calH$, if we denote $\bar X$ (resp. $\bar X'$) as the result of $m$ i.i.d. draws from $S$ (resp. $S'$) using $\mu$ (resp. $\mu'$), then, setting $\epsilon^* = \frac{6\epsilon m}{\kappa n}$ and $\delta^* = \frac{4 m e^{\epsilon^*}\delta}{\kappa n}$ we have that
\[  \Pr_{\bar X\sim \mu^m}[\calM(\bar X)\in T]\leq {e^{\epsilon^*}}  \Pr_{\bar X\sim \mu'^m}[\calM(\bar X')\in T] + \delta^*\]
\end{claim}
\begin{proof}
This follows immediately from applying Lemma 6.1 in~\cite{KarwaV18} to this setting, using the bound $\TV{\mu}{\mu'}\leq \nicefrac 1 {\kappa n}$ proven by~\citet{BunCS20}.
\end{proof}

A reader familiar with sample complexity bounds in DP-literature, knows that usually the dependency of $m$ in $\epsilon$ is inverse. Thus, the dependency of $\epsilon^*$ in $\epsilon m$ suggests that $\epsilon^*$ ends up independent of the privacy-loss of parameter set to the private learning mechanism $\calM$. That is why chose to apply $\calM$ with $\epsilon=1$, a parameter {under} which most private and non-private sample complexity bounds are asymptotically equivalent.\footnote{Loosely speaking, a parameter under which we typically get ``privacy for free.''} The flip side of it is that we now set $n$ to be proportional to $\epsilon^{-1}$.
We are now ready to give our DP algorithm for transfer learning in the hybrid model.

\begin{algorithm}[tb]
   \caption{Private Subsample-Test-Reweigh}
   \label{alg:private_transfer_learner}
\begin{algorithmic}
   \STATE {\bfseries Input:} parameters $0<\alpha,\beta<\nicefrac 1 8$, $0<\epsilon,\delta$. A $(\epsilon,\delta)$-DP learning algorithm $\calM$ of sample complexity $m(\alpha,\beta,\epsilon,\delta)$.
   \STATE \resizebox{0.48\textwidth}{!}{Set $\kappa \gets \frac{\alpha}{8(\chi^2+1)}$, $T\gets\frac{128\log_2(\frac{8(\chi^2+1)}{\alpha})}{\alpha^2}$, $N_0\gets \frac{4\ln(\nicefrac 2 \delta)\ln(\nicefrac{8T} \beta)}{\epsilon^2 \alpha^2}$.}
   \STATE Draw a sample of $n \geq m(\alpha, \frac \beta T, 1, \frac{\delta}{\epsilon\sqrt{2T}})\frac {\sqrt{288 T \ln(\nicefrac 2 \delta)}}{\epsilon \kappa}$ points i.i.d.~from $\S$, denoted $x_1, x_2, ..., x_n$, all labeled by~$\ell$. Similarly draw $N_0\cdot T$ local-users from $\T$.
   \STATE Set weight $w_i^1\gets \kappa$ for each $1\leq i \leq n$.
   \FOR {($t = 1,2,3,...,T$)}
   \STATE Set $\mu^t$ as a distribution where $\mu^t = \frac{\Pi_{\kappa}(\bar w)}{\|\Pi_\kappa(\bar w\|_1}$.
   \STATE Apply $\calM$ to a sample of $m_1=m(\alpha, \frac \beta T, 1, \frac{\delta}{\epsilon\sqrt{2T}} )$ examples drawn i.i.d.~from $\mu^t$, to obtain some hypo. $h^t$.
    \STATE Pick arbitrarily a new batch $B$ of $N_0$ local-users, and set $a^t \gets \frac{1}{N_0}\sum_{x\in B}RR_{\epsilon,\delta}\big(~\Loss(h^t(x), \ell(x)) ~\big)$. 
   \IF {($a^t > 3\alpha$)}
   \STATE $\forall i$ set $w^{t+1}_i \gets w_i^t \cdot \exp(-\nicefrac \alpha 8 \cdot [1-\Loss(h^t(x_i),\ell(x_i))])$
   \ELSE
   \STATE {\bfseries return} $h^t$ (and halt)
   \ENDIF
   \ENDFOR
\end{algorithmic}
\end{algorithm}

\begin{theorem}
\label{thm:alg_DP_boosting_is_DP}
Using the same notation as in Algorithm~\ref{alg:private_transfer_learner}, Algorithm~\ref{alg:private_transfer_learner} is a hybrid-model $(\epsilon,\delta)$-DP algorithm provided $n\geq \frac {m_1\sqrt{288 T \ln(\nicefrac 2 \delta)}}{\epsilon \kappa}= \Omega( m_1 \frac{(\chi^2+1)\sqrt{\log(\nicefrac{(\chi^2+1)} {\alpha})\log(\nicefrac 1 \delta)}}{\alpha^2 \epsilon}  )$.
\end{theorem}
\begin{proof}
First, it is clear that each local-user is asked a single query and replies using a mechanism that is $(\epsilon,\delta)$-DP. As for the privacy of the curator-agents, by setting $m_1=m(\alpha, \nicefrac \beta T, 1, \frac{\kappa \delta}{8eT} )$,  Claim~\ref{clm:sampling_into_DP} asserts that in each iteration we are $(\epsilon^*,\delta^*)$-DP for $\epsilon^* = \frac{6\cdot 1 \cdot m_1}{\kappa n}\leq \frac{\epsilon}{\sqrt{8T\ln(\nicefrac 2 \delta)}}<1$ and $\delta^* \leq \frac {4m_1e^{1}}{\kappa n}\cdot \frac{\delta}{4\epsilon \sqrt{2T}} \leq  \frac{4e\cdot\epsilon}{6\sqrt{8T}} \cdot \frac{\delta}{\epsilon \sqrt{2T}} \leq \frac{\delta}{2T}$. Applying the Advanced Composition theorem of~\cite{DworkRV10} we get that in all $T$ iterations together we are $(\epsilon, \delta)$-DP w.r.t. to each of the $n$ curator-agents.
\end{proof}

\begin{theorem}
\label{thm:alg_DP_boosting_is_useful}
W.p.$\geq 1-3\beta$, Algorithm~\ref{alg:private_transfer_learner} returns an hypothesis $h$ such that $\err_T(h)\leq 4\alpha + \alpha_{\calH}$, provided the number of curator-agents is $n=\Omega((\chi^2+1)\frac{\log(\nicefrac 1 \alpha)}{\alpha^2}(d\log(\tfrac {d} {\alpha})+\log(\tfrac 1 \beta)))$, and the number of local-agents is $N = N_0T=\Omega(\frac{\log((\chi^2+1)/\alpha)\log(\log(\chi^2+1)/\alpha\beta)}{\epsilon^2\alpha^4})$.
\end{theorem}
\begin{proof}
Again, the proof relies on the characterization of the utility of the MW-mechanism w.r.t.~$k$-dense set of weights. Again, let $\bar w^1$ be the initial weights vector where $w^1_i=\kappa$ for all $i$. Let $\bar w^*\in [0,1]^n$ be any fixed set of weights which is $\kappa$-dense, and denote $\hat w = \frac{\bar w^*}{\|\bar w^*\|_1}$ as its induced distribution. Then, we have that for any sequence of costs $\bar m^1, \bar m^2, ..., \bar m^t$, \citet{BunCS20} proved that this MW-mechanism with learning rate of $\eta = \frac \alpha 8$ guarantees that
\begin{equation}\sum_t \E_{i\sim\mu^t}[m_i^t] \leq \sum_t \E_{i \sim \hat w}[m_i^t] + \frac {\alpha T} 8 + \frac {\frac 1 {\kappa n}\phi(\bar w^*, \bar w^1)}{\alpha/8}  \label{eq:dense-MW}\end{equation}
where $\phi(\bar w^*, \bar w^1)$ is the Bregman divergence induced by the entropy function, namely 
\[ \phi(\bar w^*, \bar w^1) = \sum_i w^*_i \log(\frac{w^*_i}{w^1_i}) - w^*_i + w^1_i  \]
As $\bar w^*$ we aim to use the weights which Lemma~\ref{lem:exist_capped_weights} guarantees whose nice properties hold w.p.$\geq 1-\beta$. Thus, we set for each $x_i$ drawn from $S$ the weight $w^*_i=\frac{u(x_i)}{\nicefrac {4(\chi^2+1)}{\alpha}} \leq 1$. As Lemma~\ref{lem:exist_capped_weights} asserts, it holds that $U=\sum u(x_i)$ lies in the range $[(1 - \nicefrac{3\alpha}8)n, (1 + \nicefrac{\alpha}8)n]$, thus $w_{\rm avg} = \frac 1 n \sum \frac{\alpha u(x_i)}{{4(\chi^2+1)}} \geq  \frac{\alpha (1 - \nicefrac{3\alpha}8) }{{4(\chi^2+1)}}$. Thus, by setting $\kappa = \frac{\alpha}{{8(\chi^2+1)}} $ we have that $w_{\rm avg}\geq \kappa\cdot 1\geq \kappa w_{\max}$, implying $w^*$ is indeed $\kappa$-dense. Also, the same bounds on $U$ imply that
\begin{align*}
    \phi(\bar w^*,\bar w^1) &= \sum_i 2\kappa u(x_i)\log(\frac{2\kappa u(x_i)}{\kappa}) - 2\kappa u(x_i) + \kappa \cr
    &=\kappa \sum_i \left(2u(x_i) \log(2u(x_i)) -2u(x_i)+1\right)
    \cr &{\leq}  \kappa\big(n-2U+2\sum_i u(x_i) \log(\nicefrac{8(\chi^2+1)}{\alpha}) \big)
    \cr &\leq 2\kappa \log(\nicefrac{8(\chi^2+1)}{\alpha}) \cdot U \leq 4\kappa n  \log(\nicefrac{8(\chi^2+1)}{\alpha})
\end{align*}
The remainder of the proof is as in Theorem~\ref{thm:non_private_MW_correct}. In each iteration it must hold that $\err_{\mu^t}(h^t)\leq \alpha$; whereas $a^t\geq 3\alpha$, which w.p.$\geq 1-\beta$ ~--- using classic utility bounds for the Randomized Response mechanism~--- implies that $\err_\T(h^t)\geq 2\alpha$. Plugging this bound to Equation~\eqref{eq:dense-MW} stating the regret of the $\kappa$-dense MW-algorithm we get
\[\resizebox{0.49\textwidth}{!}{$ \sum_t (1-\alpha)]\leq \sum_t [1-(2\alpha-\frac {5\alpha} 8)] +   \frac {\alpha T} 8 + \frac{4\log_2(\frac{8(\chi^2+1)}{\alpha})}{\alpha/8} $} \]
Rearranging yields the bound
$\frac{3\alpha T }8 \leq \frac {\alpha T} 8 + \frac{32\log_2(\frac{8(\chi^2+1)}{\alpha})}{\alpha}$ which implies that we halt in $T\leq\frac{128\log_2(\frac{8(\chi^2+1)}{\alpha})}{\alpha^2}$ iterations. Again, upon halting $a^t\leq 3\alpha$ so it holds $\err_\T({h^t})\leq 4\alpha$.
\end{proof}
Thus in a realizable setting, given a finite $\calH$, we can set $\calM$ as the exponential mechanism over $\calH$ returns w.p. $\geq 1-\beta$ an hypothesis of error $\Theta(\alpha)$ when the sample size it at least $\Omega(\nicefrac {\ln(|\calH|)}{\alpha\epsilon})$. We thus obtain the following corollary.
\begin{corollary}
\label{cor:hybrid_model_learner_for_finite_realizable_H}
For any $\S,\T$ with bounded $\chi^2$-divergence, there exists a $(\epsilon,\delta)$-DP hybrid-model learning for a finite $\calH$  in the realizable case which returns w.p. $\geq 1-\beta$ an hypothesis $h\in\calH$ with $\err_\T(h)=\Theta(\alpha)$, provided that we have $n = \tilde\Omega(\frac{(\chi^2+1)\ln(|\calH|/\beta)}{\alpha^3\epsilon})$ curator-agents drawn from $\S$ and $N = \tilde\Omega(\frac{\ln(\nicefrac{(\chi^2+1)} \alpha)}{\epsilon^{2}\alpha^{4}})$ local-agents drawn from $\T$.
\end{corollary}

\paragraph{Example: Sparse Hypothses.} It is worth noting that our sample complexity bounds are independent of the dimension $d$ (assuming the domain $\calX\subset \R^d$). So consider a specific case that deals with a $s$-sparse problem over a high $d$-dimensional set where $s\ll d$, say, where $\calH$ is a linear seprartor that uses no more than $s=O(1)$ features out of the $d$ features of each example. This suggests that $|\calH|=d^{O(s)} = {\rm poly}(d)$. We thus obtain $s\cdot \rm{polylog}(d)$ sample complexity bounds (setting all other parameters to be come reasonable constants.) whereas learning solely over $\T$ requires a sample complexity $\geq d$ (see~\citet{DuchiJW13}).

\paragraph{Private SGD.} The specific case where the private learning mechanism $\calM$ is {SGD} discussed in Appendix~\ref{sec:privacy_sgd}. This case is discussed separately for two main reasons. First, its algorithmic presentation is slightly different than in Algorithm~\ref{alg:private_transfer_learner}, since in each iteration it makes successive draws from $\mu^t$ rather than a single draw of a subsample. Secondly, this is the one canonical case where $\Loss$ is continuous (rather than binary), and so the subject of scaling comes into play. Whereas thus far we assumed $\Loss$ is bounded by $1$, it is straightforward to see that a $L$-Lipfshitz convex loss function over a convex set of diameter $D$ has loss  $\in[{0}, LD]$, thus our goal is now to obtain a loss $\leq \alpha$ (rather than $\alpha \cdot LD$). Lastly, aiming to give a tight analysis, we use the notion of $(\rho,\omega)$-tCDP~\cite{BunDRS18} which is then converted to $(\epsilon,\delta)$-DP scheme. Nonetheless, the sample complexity bounds we get are similar to those of Algorithm~\ref{alg:private_transfer_learner}. 

\section{On Reducing Sample-Complexity Bounds}
\label{sec:conclusion}

While our work is the first to show the feasibility of transfer learning using poly-size sample, its sample complexity bound for the curator-agents is a large multiplicative factor $\tilde O(\epsilon^{-1}\alpha^{-2})$ over the sample complexity of a (non-transfer) curator-model learner. We believe that it is possible to significantly improve said bound, at least for particular instances. Here we provide two specific instances where indeed the number of required iterations until convergence of our algorithm is $o(\alpha^{-2})$, leading to a much smaller sample complexity.

\paragraph{Transfer Learning for {\sf PARITY} under the Uniform Distribution.} Consider the domain ${\cal X} = \{0,1\}^d$ and the class of {\sf PARITY} functions, where for any $S\subset [d]$ we have $c_S(x) = \bigoplus_{i\in S}x_i$. It is a well-known result that under the uniform distribution the {\sf PARITY} class cannot be learnt in the local-model unless the number of local-agents is $N=\exp(d)$, yet a sample of size $n=\Theta(\nicefrac {d}{\epsilon\alpha})$ suffices to learn {\sf PARITY} under any distribution in the curator-model~\cite{KasiviswanathanLNRS08}. In Appendix~\ref{apx_sec:learning_parity_unif} we show that in the hybrid-model one can learn  {\sf PARITY}  in a single iteration, provided $\S$ and the uniform distribution $\T$ have polynomial $\chi^2$-divergence. The crux of the proof lies in proving that   w.h.p.~a sufficiently large sample drawn from $\S$ is linearly independent over $\mathbb{F}_2^d$. This suggests that the private curator-model learner for {\sf PARITY}~\citep{KasiviswanathanLNRS08}~--- which leverages on (multiple) Gaussian elimination over $\mathbb{F}_2^d$~---  returns w.h.p.~the true labeling function in the {\sf PARITY} hypothesis class.

\paragraph{Empirical Experiments: When Both $\S$ and $\T$ are Gaussians.} Next, we show empirically that in other settings, both the number of required iterations until convergence and our sample complexity bounds are far greater than required. First, we consider a setting where $\S$ is a simple spherical Gaussian in $d=500$-dimensions, $\S = {\cal N}(\bar 0, I_d)$ whereas for $\T$ we picked an arbitrary set of $k=10$ coordinates and set the standard deviation on these $k$ as $\sigma = 0.02$ whereas the remaining $d-k$ coordinates have standard deviation of $1$, i.e. $\T = {\cal N}(\bar 0, I_{d-k}\otimes \sigma^2I_k)$. It is a matter of simple calculation to show that $\chisq{\T}{\S}+1 = (\frac 1 {\sigma^2(2-\sigma^2)})^{\nicefrac k 2} >3\cdot 10^{15}$. Now, our target hyperplane separator is set such that its only non-zero coordinates are the $k$ ones on which $\S$ and $\T$ have a different variance, and so that is classifies precisely $\alpha=0.01$ of the mass of $\T$ as $-1$. Now, testing this setting with the \emph{non-private version}\footnote{We thoroughly apologize, but experimenting with the private version becomes infeasible on a desktop computer due to its sample complexity constraints.} (Algorithm~\ref{alg:non_private_transfer_learner}) over $n \geq 90,000$ we obtain an hypothesis of error $\leq 2\alpha$ in all $t=50$ repetitions of our experiment. Moreover, our iterative algorithm runs for only $T\approx 1200-1300$ iterations, far below the $\approx \alpha^{-2}$ upper bound. The results appear in Figures~\ref{fig:error_lin} and~\ref{fig:iterations_lin} in Appendix~\ref{apx_sec:experiments}. Second, we consider a setting where the true hypothesis actually corresponds to a $k$-dimensional ball over the $k$ coordinates that are more concentrated in $\T$ than in $\S$ and ran both the private and non-private version of our algorithm. While the non-private version converged very fast, the private version required a very large sample of curator-agents, and so we were able to conduct only preliminary experiments with it. The experiment is detailed in Appendix~\ref{apx_sec:experiments_bounded} where the results appear in Figures~\ref{fig:all_results_noDP} and~\ref{fig:all_results_DP}.

\paragraph{Open Problems.} 
Our work is the first to present a general framework for transfer learning in the hybrid-model, thus providing an initial answer to the open problem of ``selection bias'' posed by~\citet{BeimelKNSS20}. While our framework does surpasses certain na\"ive baselines in some specific cases, there is still much work to be done in order to reduce its sample complexity. Considering different divergences can be a promising direction, especially if we know that the $4$th moment of the IS weights is bounded (as it may increase the value of $\kappa$). A different venue can be the study of repeated uses of Subsample-Test-Reweigh~--- when person A applied the paradigm until she finds a good hypothesis, then hands over her last set of weights to Person B who uses it for learning a different hypothesis. Lastly, we believe that there's more to be studied in general in the intersection between DP and transfer learning, as the problem can be tackled from the alternative approach of \emph{discrepancy} between hypothesis classes~\cite{ben2010theory,mansour2009domain}.

\section*{Acknowledgements}

This work was done when the first author was advised by the second author. O.S.~is supported by the BIU Center for Research in Applied Cryptography and Cyber Security in conjunction with the Israel National Cyber Bureau in the Prime Minister's Office, and by ISF grant no.~2559/20. Both authors thank the anonymous reviewers for many helpful suggestions in improving this paper.

\bibliography{paper}

\appendix

\newpage
\ 
\newpage
\section{Private Stochastic Gradient Descent}
\label{sec:privacy_sgd}

\paragraph{The Private SGD Algorithm.}
In this section, we discuss our private Subsample-Test-Reweigh paradigm where the private learning mechanism applied in each iteration is the standard private SGD (see~\citet{Hazan16,BassilyST14}). For simplicity, we give the algorithm here.

\begin{algorithm}
   \caption{Online SGD}
   \label{alg:private_sgd}
\begin{algorithmic}
   \STATE {\bfseries Input:} Parameters $\alpha,\beta,\sigma$ Lipfshitz constant $L$ and a convex set $\calH\subset \R^d$ of diameter $D$. A distribution $\mu$ over a sample $S$.
   \STATE Let $h^1 \in \calH$ arbitrarily.  
   \STATE Set $R\gets\max\big\{  \frac{9D^2(L^2+\sigma^2d)}{\alpha^2}, \frac{8D^2L^2\ln(\nicefrac 2 \beta)}{\alpha^2}  \big\}$; or set
   $R\gets \max\big\{  \frac{4(L^2+\sigma^2d)}{\lambda\alpha}\ln(\frac{L^2+\sigma^2d}{\lambda\alpha}), \frac{8D^2L^2\ln(\nicefrac 2 \beta)}{\alpha^2}  \big\}$ if $\Loss$ is $\lambda$-strongly convex.
   \FOR {($r = 1,2,3,...,R$)}
   \STATE Draw a labeled example $(x_r, \ell(x_r)) \sim \mu$
   \STATE Draw a random vector $v\sim {\cal N}(0, \sigma^2 I_d)$
   \STATE Set $\eta^r \gets \frac{D}{\sigma \sqrt r}$; or set $\eta^r \gets \frac{D}{\lambda r}$ if $\Loss$ is $\lambda$-strongly convex.
   \STATE Set $h^{r+1} \gets \Pi_{\calH} \big(h^r -\eta^r( \nabla \Loss( h^r(x_r), \ell(x_r)) + v)\big)$
   \ENDFOR
   \STATE
   \textbf{Return} $\bar{h} = \frac{1}{R} \sum_{r=1}^R h^r$.
\end{algorithmic}
\end{algorithm}

Standard arguments from~\citet{Hazan16} give the following utility theorem.
\begin{theorem}
\label{thm:utility_private_SGD}
Let $\calH\subset \R^d$ be a convex set of diameter $D$ and let $\Loss$ be a $L$-Lipfshitz function and denote $\alpha_{\calH} = \min_{h\in\calH}\E_{(x,\ell(x))\sim \mu}[\Loss(h(x),\ell(x))]$. Then, w.p. $\geq 1-\beta$ after $R$ iterations, $\err(\bar h)\leq \alpha_{\calH}+\alpha$.
\end{theorem}
\begin{proof}
The proof follows from the usual analysis of SGD, under the observation that in each iterations
\begin{align*} 
&\E\nolimits_{\substack{(x_r,\ell(x_r))\sim \mu\\v\sim{\cal N}(0,\sigma^2I_d)}}[\nabla \Loss( h^r(x_r), \ell(x_r)) + v]\cr & \qquad \qquad \qquad \qquad = \E\nolimits_{(x,y)\sim \mu}[\nabla\Loss(h^r(x),\ell(x))]
\intertext{and that}
 &\E[\|\nabla \Loss( h^r(x_r), \ell(x_r)) + v\|^2] \leq \E[\|\nabla \Loss( h^r(x_r), \ell(x_r))\|^2]
\cr & ~~~~+ 2\E[\nabla \Loss( h^r(x_r), \ell(x_r))\cdot v]+\E[\|v\|^2]\leq L^2 + \sigma^2d\end{align*} 
Plugging those into the bounds of the generalization properties of the SGD for convex functions we obtain:
\[ \err_\mu(\bar h) \leq  \err_\mu(h^*)  +  \frac{3D\sqrt{(L^2+d\sigma^2)}}{2\sqrt{R}} + LD\sqrt{\frac{8 \ln(\nicefrac{2}{\beta})}{R}} \]
So setting $R \geq \max\big\{  \frac{9D^2(L^2+\sigma^2d)}{\alpha^2}, \frac{8D^2L^2\ln(\nicefrac 2 \beta)}{\alpha^2}  \big\}$ yields that $\err_\mu(\bar h)\leq \alpha_{\calH}+\frac \alpha 2 + \frac \alpha 2$ as required. Similarly, for a $\lambda$-strongly convex function we get
\[ \err_\mu(\bar h) \leq  \err_\mu(h^*)  +  \frac{(L^2+d\sigma^2)(1+\ln(T))}{2\lambda R} + LD\sqrt{\frac{8 \ln(\nicefrac{2}{\beta})}{R}} \]
So setting $R \geq \max\big\{  \frac{4(L^2+\sigma^2d)}{\lambda\alpha}\ln(\frac{L^2+\sigma^2d}{\lambda\alpha}), \frac{8D^2L^2\ln(\nicefrac 2 \beta)}{\alpha^2}  \big\}$ yields that $\err_\mu(\bar h)\leq \alpha_{\calH}+\alpha$ as required.
\end{proof}

While we can apply a privacy analysis of Algorithm~\ref{alg:private_sgd}, we table it to the privacy analysis of our full algorithm.

\paragraph{The Subsample-Test-Reweigh Using SGD.} We now transition to the full analysis of STR when we apply SGD as an intermediate procedure. 

\begin{algorithm}[tb]
   \caption{Private Subsample-Test-Reweigh with SGD}
   \label{alg:private_STR_with_SGD}
\begin{algorithmic}
   \STATE {\bfseries Input:} parameters $0<\alpha,\beta<\nicefrac 1 8$, $0<\epsilon<1$,$0<\delta<\nicefrac 1 4$. Private SGD mechanism over a convex set $\calH\in \R^d$ of diameter $D$ with convex $L$-Lipfshitz loss function $\Loss$ making at most $R$ SGD-iterations.
   \STATE
   \STATE Set $\kappa \gets \frac{\alpha}{8(\chi^2+1)}$,  $T\gets\frac{128L^2D^2\log_2(\frac{8(\chi^2+1)}{\alpha})}{\alpha^2},  N_0\gets\linebreak \frac{4\ln(\nicefrac 2 \delta)\ln(\nicefrac{8T} \beta)}{\epsilon^2 \alpha^2}, {\nolinebreak\sigma^2 \gets \max\big\{ 20L^2, \frac{16\epsilon^{-1}L^2\ln(\nicefrac 1 \delta)+8L^2}{\ln(\kappa n)}\big\}}$.
   \STATE Set $R\gets\max\big\{  \frac{9D^2(L^2+\sigma^2d)}{\alpha^2}, \frac{8D^2L^2\ln(\nicefrac 2 \beta)}{\alpha^2}  \big\}$; or set
   $R\gets \max\big\{  \frac{4(L^2+\sigma^2d)}{\lambda\alpha}\ln(\frac{L^2+\sigma^2d}{\lambda\alpha}), \frac{8D^2L^2\ln(\nicefrac 2 \beta)}{\alpha^2}  \big\}$ if $\Loss$ is $\lambda$-strongly convex.
   \STATE Draw a sample of $n \geq \sqrt{ \frac{52RT}{8\kappa^2\epsilon } \ln(\frac{52RT}{8\kappa^2\epsilon }) }$ points i.i.d.~from $\S$, denoted $x_1, x_2, ..., x_n$, all labeled by~$\ell$. 
   \STATE Draw $N_0\cdot T$ local-users from $\T$.
   \STATE Set weight $w_i^1\gets \kappa$ for each $1\leq i \leq n$.
   \FOR {($t = 1,2,3,...,T$)}
   \STATE Set $\mu^t$ as a distribution where $\mu^t = \frac{\Pi_{\kappa}(\bar w)}{\|\Pi_\kappa(\bar w\|_1}$.
   \STATE Apply private SGD using $\mu^t$ and using $\sigma^2$ for $R$ iterations, and obtain an hypothesis $h^t$.
    \STATE Pick arbitrarily a new batch $B$ of $N_0$ local-users, and set $a^t \gets \frac{1}{N_0}\sum_{x\in B}RR_{\epsilon,\delta}\big(~\Loss(h^t(x), \ell(x)) ~\big)$. 
   \IF {($a^t > 3\alpha +\alpha_\calH$)}
   \STATE $\forall i$ set \break $w^{t+1}_i \gets w_i^t \cdot \exp(-\nicefrac \alpha {8LD} \cdot [LD-\Loss(h^t(x_i),\ell(x_i))])$
   \ELSE
   \STATE {\bfseries return} $h^t$ (and halt)
   \ENDIF
   \ENDFOR
\end{algorithmic}
\end{algorithm}

The utility analysis of Algorithm~\ref{alg:private_STR_with_SGD} is just as the one of Algorithm~\ref{alg:private_transfer_learner} modulo the fact that the non-negative loss is now bounded by $LD$ rather than $1$, which implies we must increase the number of MW-iterations by $L^2D^2$. It requires a sample complexity of  $n=\Omega(L^2D^2(\chi^2+1)\frac{\log(\nicefrac 1 \alpha)}{\alpha^2}(d\log(\tfrac {d} {\alpha})+\log(\tfrac 1 \beta)))$ curator-agents, and $N = N_0T=\Omega(\frac{L^2D^2\log((\chi^2+1)/\alpha)\log(\log(\chi^2+1)/\alpha\beta)}{\epsilon^2\alpha^4})$ local-agents to return, w.p.$\geq 1- 3\beta$ an hypothesis of error $\leq \alpha_{\calH}+\alpha$. (The increase in the number of curator-agents is due to the fact that the loss is now in the range $[0,LD]$ rather than $[0,1]$.) The privacy analysis of Algorithm~\ref{alg:private_STR_with_SGD} requires we transition of $(\rho,\omega)$-tCDP given in~\citet{BunDRS18}. Its definition as well as some of its basic properties (proven in~\citet{BunS16, Mironov17, BunDRS18}) are provided below.

\begin{definition}
\label{def:tCDP}
A mechanism $\calM$ is said to be $(\rho,\omega)$-tCDP if for two neighboring inputs $I $ and $I'$ the $\alpha$-Reyni divergence of the two distributions is bounded: $D_{\alpha}(\calM(I)||\calM(I'))\leq \alpha \rho$ for any $1<\alpha\leq \omega$.
\end{definition}
\begin{fact}
\label{fast:gaussian_mechanism}
\begin{itemize}
    \item 
Let $f$ be a $d$-dimensional function of $L_2$-global sensitivity of $\max_{I,I' \textrm{ neighbors}}\|f(I)-f(I')\|\leq L$. Then the mechanism that outputs for any instance $I$ an output drawn from ${\cal N}(f(I), \sigma^2 I_d)$ is $(\frac{2L^2}{\sigma^2},\infty)$-tCDP.
    \item Let $\calM_1, \calM_2$ be two $(\rho_1,\omega_1)$-tCDP (resp. $(\rho_2,\omega_2)$-tCDP) mechanisms. Then the mechanism that applies them to the same instance but using independent coin toss for each is $(\rho_1+\rho_2, \min\{\omega_1, \omega_2\})$-tCDP.
    \item A mechanism which is $(\rho,\omega)$-tCDP is also \linebreak $(\rho\omega+\frac{\ln(\nicefrac 1 \delta)}{\omega-1},\delta)$-DP for any $\delta < \nicefrac 1 e$.
\end{itemize}
\end{fact}

Perhaps, however, the most important property of tCDP is that it is amplified by subsampling. 
\begin{theorem}
\label{thm:tCDP_subsampling}[Thm.~12 of~\citet{BunDRS18} reworded] Fix $\rho \in (0, 0.1]$. Let $s$ be a constant satisfying $\log(\nicefrac 1 s)\geq 3\rho(2+\ln(\nicefrac 1 \rho))$. Let $\calM$ be a $(\rho,\infty)$-tCDP mechanism. Let $I$ and $I'$ be two neighboring instance and let $\mu$ be a distribution where the probability of sampling the one different entry between the two instances is $s$. Then the mechanism that samples entries from the input and then applies $\calM$ on the subsample is $(13s^2\rho, \frac{\log(\nicefrac 1 s)}{4\rho})$-tCDP.
\end{theorem}

Now, throughout the execution of Algorithm~\ref{alg:private_STR_with_SGD} it holds that we subsample a point and apply the Gaussian mechanism for $RT$ iterations. In all of these iterations we apply a distribution where the probability of subsampling any point into $\calM$ is at most $\nicefrac 1 {\kappa n}$. Moreover, our private-SGD mechanism when applied to a $L$-lipfshitz loss function is $(\frac{2L^2}{\sigma^2},\infty)$-tCDP. Thus, if $\frac{2L^2}{\sigma^2}\leq 0.1$ and if 
\begin{equation}
    \label{eq:subsampling_constraint}
    \ln(\kappa n)\geq \frac{6L^2 (2+\ln(\nicefrac {\sigma^2} {2L^2}))}{\sigma^2}
\end{equation}
then all conditions of Theorem~\ref{thm:tCDP_subsampling} hold. This suggests that each time we execute $\calM$ over a randomly drawn sample we are $(\frac{26L^2}{\sigma^2\kappa^2n^2}, \frac{\sigma^2\ln(\kappa n)}{8L^2})$-tCDP; thus, by composition, we are $(\frac{26L^2 RT}{\sigma^2\kappa^2n^2}, \frac{\sigma^2\ln(\kappa n)}{8L^2})$-tCDP. Thus, w.r.t.~the curator-agents, we are $(\epsilon,\delta)$-DP for any $\delta<\nicefrac 1 4$ for
\[\epsilon = \frac{26 RT\ln(\kappa n)}{8\kappa^2n^2} + \frac{8L^2\ln(\nicefrac 1 \delta)}{\sigma^2\ln(\kappa n)-8L^2}  \] 

This suggests that in order to achieve $(\epsilon,\delta)$-DP w.r.t. the curator agents, we set $\sigma^2 = \frac{16\epsilon^{-1}L^2\ln(\nicefrac 1 \delta)+8L^2}{\ln(\kappa n)}$, and we must set $n$ so that $n \geq \sqrt{\frac{52RT[\ln(\kappa)+\ln(n)]}{8\kappa^2\epsilon }}$. Seeing as $\kappa <1$ it thus suffices for us to set $n = \sqrt{ \frac{52RT}{8\kappa^2\epsilon } \ln(\frac{52RT}{8\kappa^2\epsilon }) }$. It remains to check that under these values~\eqref{eq:subsampling_constraint} holds; but indeed, note that $\frac{\sigma^2}{2L^2}\geq 10$ and so
\begin{align*}
&\frac{3(2+\ln(\nicefrac {\sigma^2} {2L^2}))}{\nicefrac {\sigma^2}{2L^2}} \leq \frac{3(2+\ln(10))}{10} < 1.3 
\end{align*}
whereas under any reasonable set of parameters we get that $n\geq \sqrt{ \frac{52RT}{8\kappa^2\epsilon } \ln(\frac{52RT}{8\kappa^2\epsilon }) } >\frac 4 \kappa$, implying that $\ln(\kappa n)\geq \ln(4)>1.3$.

\begin{theorem}
\label{thm:STR_with_SGD_is_private} For any given $\epsilon \in (0,1)$, $\delta \in (0,\nicefrac 1 4)$, $0<\alpha<\nicefrac 1 8$, $0<\beta<\nicefrac 1 3$, two distribution $\S,\T$ with bounded $\chi^2$-divergence and a convex loss-function $\Loss$ over a convex set $\calH\subset \R^d$ of diameter $D$, we have that Algorithm~\ref{alg:private_STR_with_SGD} is $(\epsilon,\delta)$-DP algorithm in the hybrid model provided that
\begin{align*}n &= \tilde \Omega( \frac{(\chi^2+1)LD^2\sqrt{L^2+\sigma^2 d}\cdot  \sqrt{\ln(\nicefrac 2 \beta)}} {\alpha^3 \sqrt\epsilon}  )
\cr &= \tilde \Omega( \frac{(\chi^2+1)D^2L^2\sqrt{d\ln(\nicefrac 1 \delta)}\cdot  \sqrt{\ln(\nicefrac 2 \beta)}} {\alpha^3 \epsilon}  )
\end{align*}
if the loss-function is $L$-Lipfshitz and convex, or provided that 
\begin{align*}
     n &= \tilde \Omega( \frac{LD(\chi^2+1)\sqrt{\frac{L^2+\sigma^2 d}{\lambda\alpha}+\frac{D^2L^2\ln(\nicefrac 1 \beta)}{\alpha^2}}} {\alpha^2  \sqrt{\epsilon}}  )
     \cr &= \tilde\Omega( \frac{LD(\chi^2+1)}   {\alpha^2} \cdot (\frac{L\sqrt{d\ln(\nicefrac 1 \delta)}}{\epsilon\sqrt{\alpha\lambda}}+\frac{DL\sqrt{\ln(\nicefrac 1 \beta)}}{\sqrt{\epsilon}\alpha}))
\end{align*}
if the loss-function is $L$-Lipfshitz and $\lambda$-strongly convex. Furthermore, if the number of local-agents is $N =\Omega(\frac{L^2D^2\log((\chi^2+1)/\alpha)\log(\log(\chi^2+1)/\alpha\beta)}{\epsilon^2\alpha^4})$ then w.p. $\geq 1 - 3\beta$ it returns an hypothesis $h\in \calH$ where $\err_\T(h)\leq \alpha_{\calH}+4\alpha$.
\end{theorem}

\section{Transfer Learning for the {\sf PARITY}-Problem Under the Uniform Distribution}
\label{apx_sec:learning_parity_unif}

\paragraph{Transfer Learning of {\sf PARITY} for the uniform distribution.} Consider the domain ${\cal X} = \{0,1\}^d$ and the class of {\sf PARITY} functions, where for any $S\subset [d]$ we have $c_S(x) = \bigoplus\limits_{i\in S}x_i$. It is a well-known result that under the uniform distribution the {\sf PARITY} class cannot be learnt in the local-model unless the number of local-agents is $N=\exp(d)$, yet a sample of size $n=\Theta(\nicefrac {d}{\epsilon\alpha})$ suffices to learn {\sf PARITY} under any distribution in the curator-model~\cite{KasiviswanathanLNRS08}. Here we show that in the hybrid-model one can learn the {\sf PARITY} class with a single iteration, provided $\S$ and the uniform distribution $\T$ have polynomial $\chi^2$-divergence.

To establish this, we prove the following sequence of  claims and corollaries. 
\begin{proposition}
\label{pro:independent_set_uniform_dist} Fix $0<\beta<\nicefrac 1 2$. Let $S$ be a sample of $n\geq d\log_2(\nicefrac d \beta)$ points drawn i.i.d.~from the uniform distribution over $\{0,1\}^d$. Then the probability that $S$ isn't linearly independent is $\leq \beta$.
\end{proposition}
\begin{proof}
We prove the claim inductively as we iterate over all vectors in $S$. Due to simple counting argument, it is easy to see that the probability to draw an vector that is linearly dependent of given set of $i$ vectors in a the $d$-dimension space $\mathbb{F}_2^d$ is at most $2^{i-d}$. So fix any $0\leq i \leq d-1$. At each step we have a set of $i$ linearly independent vectors spanning a subspace of dimension $i$. We argue that the probability that among the next $t=\log_2(d/\beta)$  vectors in $S$ the probability that all $t$ vectors are linearly dependent of these $i$ basis vectors is at most $(2^{i-d})^t \leq 2^{-t} = \nicefrac \beta d$. And so, after $d$ iterations we have found a set of $d$ linearly independent vectors in $S$ w.p. $\geq 1-\beta$.
\end{proof}
\begin{claim} 
\label{clm:independent_set_non_uniform_dist} Fix $0<\beta<\nicefrac 1 2$. 
Let $\T$ be the uniform distribution over $\{0,1\}^d$ and let $\S$ be a distribution over the same domain s.t. $\chisq{\T}{\S}=\chi^2$ is finite. Let $S$ be a sample of $n\geq 4(\chi^2+1)d\ln(\nicefrac d \beta)$ points drawn i.i.d.~from $\S$. Then the probability that $S$ isn't linearly independent is $\leq \beta$.
\end{claim}
\begin{proof}
The claim is proven in a similar inductive fashion to Proposition~\ref{pro:independent_set_uniform_dist}. Fix any $0\leq i \leq d-1$. At each step we have a set of $i$ linearly independent vectors spanning a subspace of dimension $i$. We argue that the probability that among the next $t=\log_2(d/\beta)$  vectors in $S$ the probability that all $t$ vectors are linearly dependent of these $i$ basis vectors is at most $\leq \nicefrac \beta d$, from which the claim follows immediately.

So now, given $i$ linearly independent vectors, let $E$ be the event that we draw a vector not in their span. Under the uniform distribution $\Pr_{\T}[E] \geq 1-2^{i-d}\geq \nicefrac 1 2$. Standard bounds on the $\chi^2$-divergence give that
\[ \sqrt{\Pr_{\S}[E](\chi^2+1)}\geq \Pr_{\T}[E]\geq \nicefrac 1 2 \] 
implying that $\Pr_{\S}[E]\geq \frac 1 {4(\chi^2+1)}$ and so $\Pr_\S[\bar E]\leq 1-\frac 1 {4(\chi^2+1)}$. It follows that the probability that among the next $t = 4(\chi^2+1)\ln(\nicefrac d \beta)$ draws, not a single one lies outside the span of these $i$ vectors is at most ${(1-\frac 1 {4(\chi^2+1)})^t \leq} \exp(-\frac {t} {4(\chi^2+1)})=\nicefrac \beta d$ as required.
\end{proof}

\begin{corollary}
\label{cor:parity_learnable_chi_sq} Fix $\beta>0$. Set $k = \log_{\nicefrac 4 3}(\nicefrac 2 \beta)$. Under the same notation as in Claim~\ref{clm:independent_set_non_uniform_dist} let $S_1, S_2, ..., S_k$ be $k$ independently drawn batches from $\S$ s.t. each $S_i$ contains at least $n\geq \frac{32(\chi^2+1)d\ln(\nicefrac {2dk} \beta)}{\epsilon}$. Then w.p. $\geq 1-\beta$ it holds that when we drawn from each $S_i$ a subsample $S'_i$ where each $x\in S_i$ is put in the subsample w.p. $\nicefrac \epsilon 4$ independently of all other examples, then all $S_i'$ are linearly independent.
\end{corollary}
\begin{proof}
Straight-forward application of the Chernoff bound gives that w.p.$\geq 1-\nicefrac \beta 2$ each of the $k$ $S'_i$-s contains at least $\nicefrac \epsilon 8 |S_i|$ many points. This suffices for us to apply Claim~\ref{clm:independent_set_non_uniform_dist} and have that w.p. $\geq 1 -\nicefrac \beta 2$ all $S'_i$ are linearly independent.
\end{proof}
Based on Corollary~\ref{cor:parity_learnable_chi_sq} we can now apply the same {\sf PARITY} learning algorithm from~\citet{KasiviswanathanLNRS08} on the $kn$ curator-agents \emph{just once} and then test its correctness over the $N$. This algorithm outputs for each $S_i'$ either a $\bot$ or a solution in the affine subspace that solves a system of equations over $\mathbb{F}_2$. But due to the linear independence of each $S_i'$, this solution must be the indicating vector of the relevant features of the true classifying function $c_S^*\in {\sf PARITY}$. It follows that for each $S_i$ outputs the true classifying function w.p.$\geq 1/4$; and so, w.p. $\geq 1-\beta$ all $S_i$ return either $\bot$ or $c_S^*$ where at least one of the $S_i$-s outputs the true classifier. Note that the true classifier's loss -- under any distribution $\S$ or $\T$ -- must be $0$. Using additional $O(\epsilon^{-2}\alpha^{-2})$ we can test and see that indeed we have an hypothesis $c\in {\sf PARTIY}$ which is of small loss.

\section{Experimental Evaluation of Non-Private Subsample-Test-Reweigh.}
\label{apx_sec:experiments}

In this section, we show empirically that in another setting, both the number of required iterations until convergence and our sample complexity bounds are far greater than required. We consider a setting where $\S$ is a simple spherical Gaussian in $d=500$-dimensions, $\S = {\cal N}(\bar 0, I_d)$ whereas for $\T$ we picked an arbitrary set of $k=10$ coordinates and set the standard deviation on these $k$ as $\sigma = 0.02$ whereas the remaining $d-k$ coordinates have standard deviation of $1$, i.e. $\T = {\cal N}(\bar 0, I_{d-k}\otimes (0.01)^2I_k)$. It is a matter of simple calculation to show that $\chisq{\T}{\S}+1 = (\frac 1 {\sigma^2(2-\sigma^2)})^{k/2}> 3\cdot 10^{15}$. Now, true hypothesis is a hyperplane separator set on the $k$ coordinates  on which $\S$ and $\T$ have a different variance, so that is classifies precisely $\alpha=0.01$ of the mass of $\T$ as $-1$.

\begin{figure}
\begin{subfigure}[t]{.49\textwidth}
    \centering
    \includegraphics[scale=0.45]{"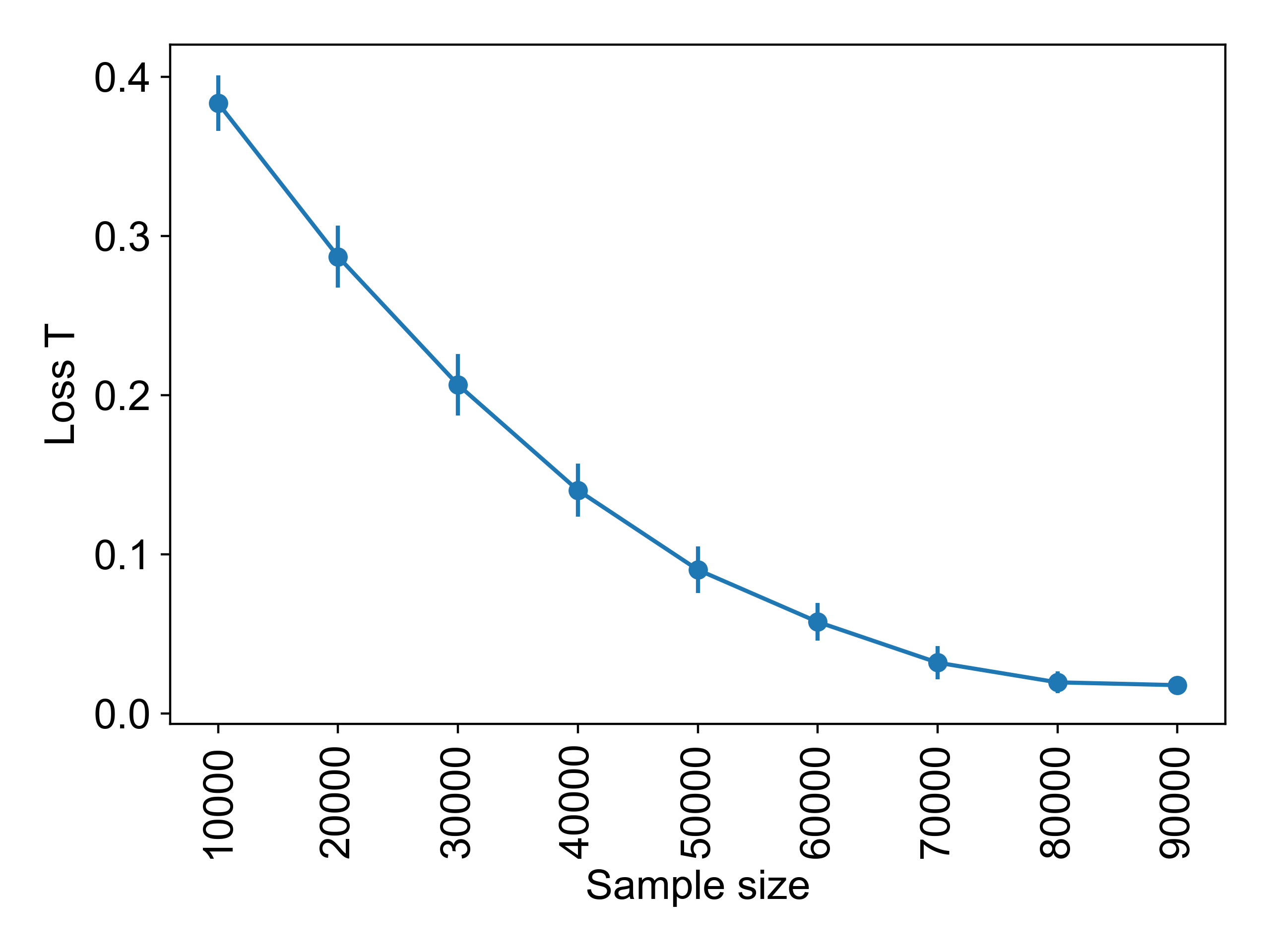"}
    \caption{\label{fig:error_lin}\textbf{Loss on $\T$}: The loss on distribution $\T$ until convergence (in samples 90K-140K) or (in samples 10K-80K) until arriving to early stopping condition (the average loss on T in last 200 iterations not improved in more 0.01  compared to the best average loss in the previous 200 iterations).}
  \end{subfigure}
  \begin{subfigure}[t]{.49\textwidth}
    \centering
    \includegraphics[scale=0.45]{"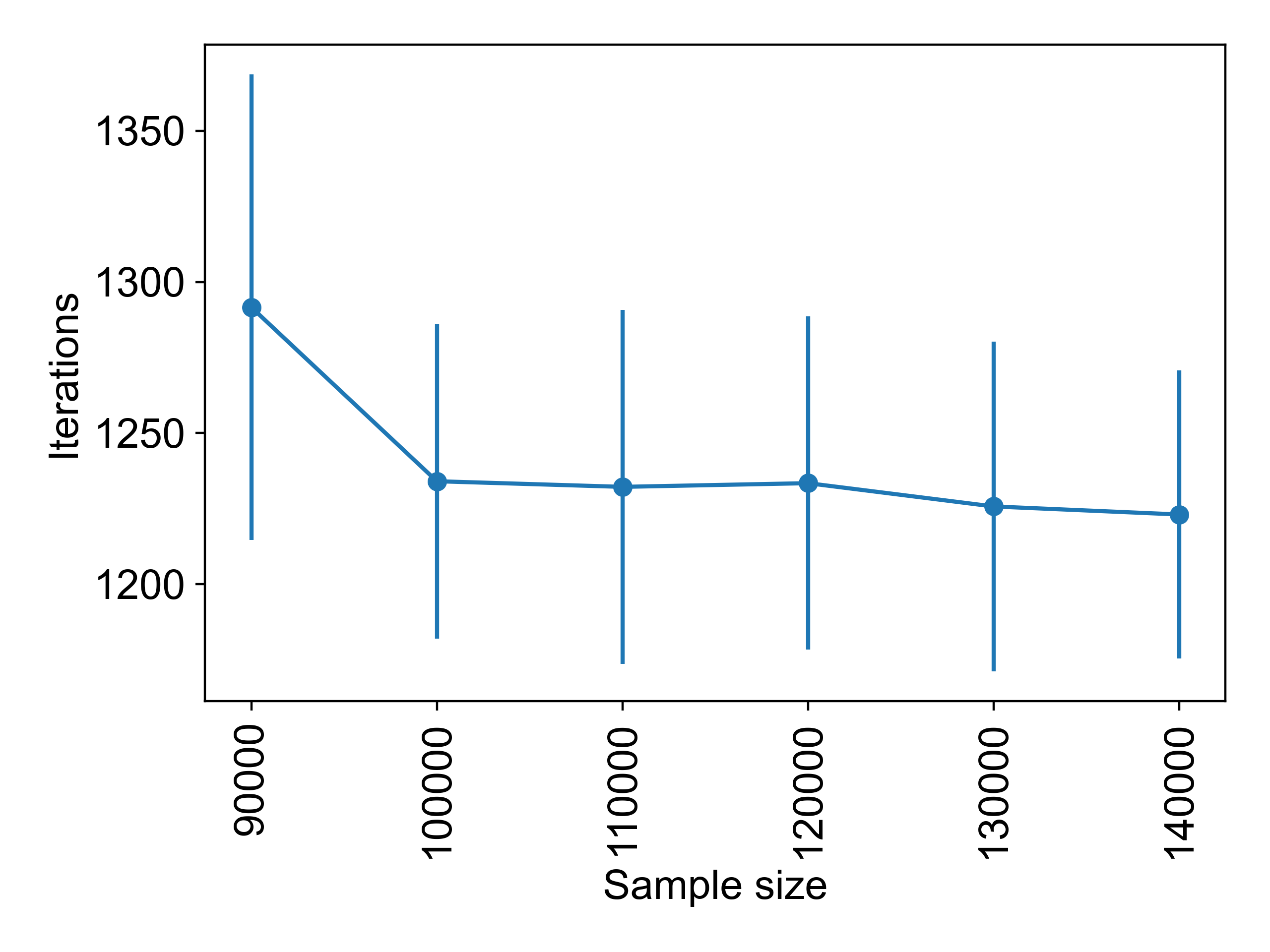"}
    \caption{\label{fig:iterations_lin}\textbf{Iterations number}: The iterations number until convergence decreases with size of the sample.}
  \end{subfigure}
  \begin{subfigure}[t]{.49\textwidth}
    \centering
    \includegraphics[scale=0.5]{"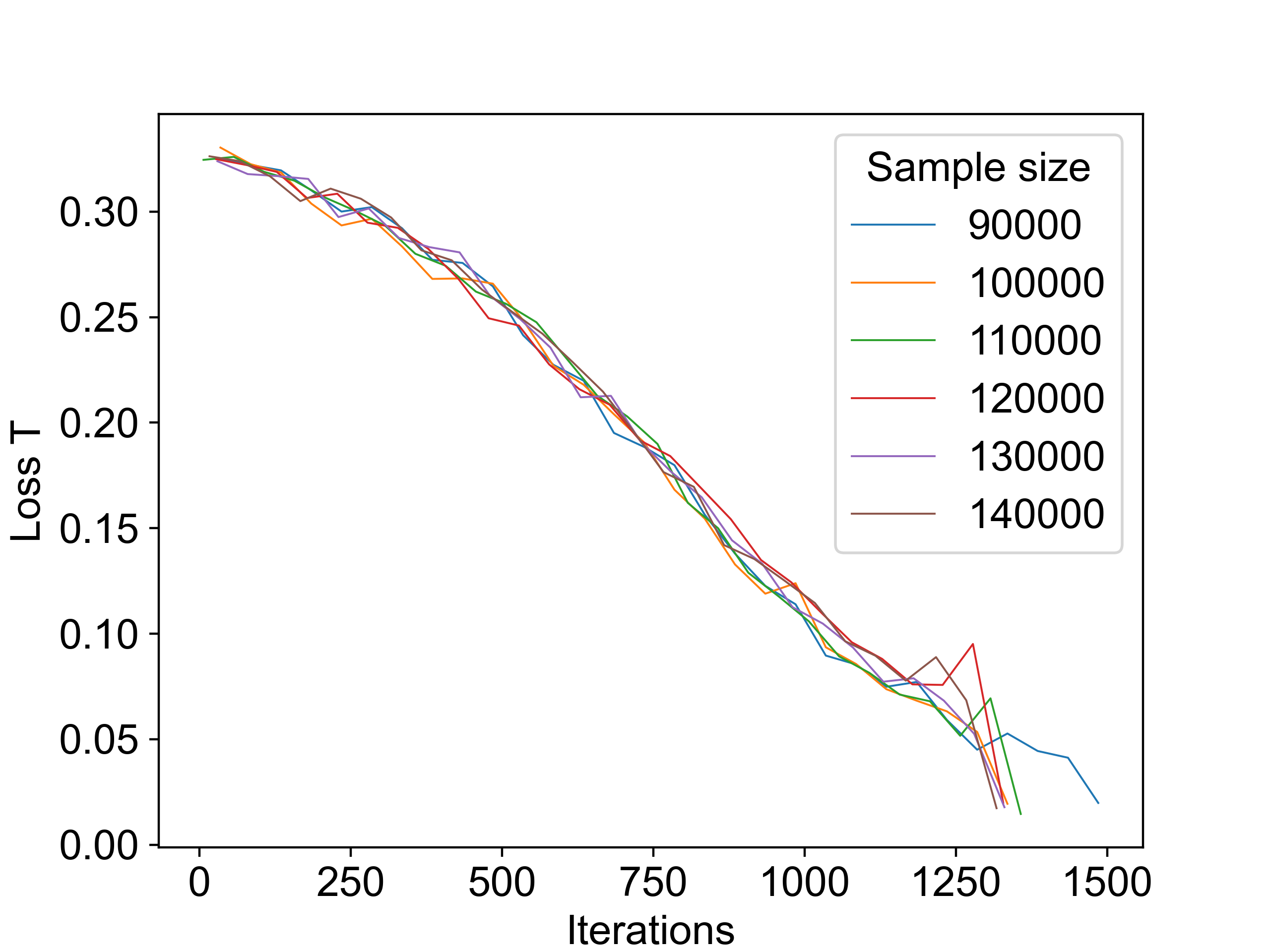"}
    \caption{\label{fig:error_on_run_lin}\textbf{Loss on $\T$ along the run}: the loss on $T$ decreases with the iterations.}
  \end{subfigure}
  \caption{\label{fig:all_results_lin} Empirical Experiment Results}
\end{figure}

We applied the non-private version of our algorithm (Algorithm~\ref{alg:non_private_transfer_learner}). In the non-private version, in order to learn in each iteration a hyperplane separator over a subsample of examples from $\S$ we used SVM, where our optimization goal is $\tfrac 1 2\|w\|^2 + C\cdot \sum_x \max\{0, 1-\langle w,x \rangle\}$ with a very large $C=10^{30}$ (aiming to find an exact hyperplane as possible) over a subsample whose size is set to be $\frac{d+\ln(0.05/T)} \alpha$. First, aiming to find the sample complexity of the curator-agents that already yields an hypothesis of error $\leq 2\alpha$, we ran our experiments with varying values of $n = \{ 10^4, 2\cdot 10^4, ..., 8\cdot 10^4, 9\cdot 10^4 \}$, repeating each experiment $t=50$ times. We observed that for $n=80,000$ we consistently return an hypothesis of small-loss. Results appear in Figure~\ref{fig:error_lin}. Then, for even larger values of $n = \{9\cdot 10^4, 1\cdot 10^5, .., 1.4\cdot 10^5\}$ we ran our experiment to see at which iteration do we halt. For $n=80,000$ we halt after $\approx 1300$ iterations whereas for $n=140,000$ we halt after $T\approx 1200$; yielding the rather surprising result that $T$ isn't greatly affected by the increasing sample complexity. But regardless, this is very far below than the $O(\alpha^{-2})$-upper bound. Results appear in Figure~\ref{fig:iterations_lin}. In fact, looking at the error of the resulting hypothesis along the run itself returns roughly the same values, as seen in Figure~\ref{fig:error_on_run_lin}.


\section{Experimental Evaluation of Private Subsample-Test-Reweigh}
\label{apx_sec:experiments_bounded}

In order to implement the private algorithm, we used another example with bounded examples and hypotheses. 

Similarly, to the previous experiment, we consider a setting where $\S$ is a simple spherical Gaussian in $d=200$-dimensions, $\S = {\cal N}(\bar 0, I_d)$ whereas for $\T$ we picked an arbitrary set of $k=6$ coordinates and set the standard deviation on these $k$ as $\sigma = 0.4$ whereas the remaining $d-k$ coordinates have standard deviation of $1$, i.e. $\T = {\cal N}(\bar 0, I_{d-k}\otimes (0.4)^2I_k)$. It is a matter of simple calculation to show that $\chisq{\T}{\S}+1 = (\frac 1 {\sigma^2(2-\sigma^2)})^{k/2}> 39.1$. 

However, we now set the true labeling function as one that \emph{correlated} to points with large importance sampling weight. It is fairly simple to see that by looking at the $k$-coordinates with smaller variance in $\T$, any origin-centered ball has more probability mass in $\T$ then in $\S$. And so, our hypothesis class is the set of origin-centered ellipses $\calH = \{w_1, ..., w_d,b \in [0,1]: ~\sum_{i=1}^d w_i x_i^2  \le b\}$ where so that $x_i \sim \S$ and thus $\bar{y} \sim \chi_k^2$. The true hypothesis $\ell$ is the hyperplane separator with  $\bar{w} = 0 \cdot I_{d-k}\otimes 1\cdot I_k$ and $b=\sigma^2 r^0$, where $r^0$ is the a numerically set threshold under which a $\Pr_{X\sim \chi^2_k}[X < r^0]=0.3$. Thus $\ell$ labels precisely $30\%$ of the probability mass of $\T$ as $-1$, whereas it labels a significantly smaller fraction of $\S$ as $-1$.

As the curator-model learning algorithm we used the online SGD as presented in~\citet{Hazan16}, after mapping each example $x\in \R^d$ to the vector $y\in \R^d_+$ where $y_i=x_i^2$ for each coordinate $y$. This allows us to use the Hinge-loss function ~- $\Loss(y)=\max\{0, \ell^*(y)(\langle w,y \rangle -b)\}$. The learning rate of the algorithm depends on a bounded diameter of the hypothesis and Lipschitzness of the loss that upper-bound the $\|\bar{y}\|$. So we set a value $B=27.9$ -  the empirically the max value of $\|\bar{y}\|$ of $90\%$ of the examples drawn from $\S$, and projected each example with norm $>B$ onto this simplex: $\{y\in \R^d_+:~ \|y\|\leq B\}$. Therefore, including the additional intercept coordinate, we get: $\| \nabla \Loss(\bar{y}) \|^2 \leq B^2 +1$, so we can use the Lipschitz parameter of $L = \sqrt{B^2 +1}$. We also verify that the diameter is lower than $\sqrt{d+1}$ by verifying that $\sigma \le \nicefrac{1}{\sqrt{r}}$.

\begin{figure}
\begin{subfigure}[t]{.49\textwidth}
    \centering
    \includegraphics[scale=0.45]{"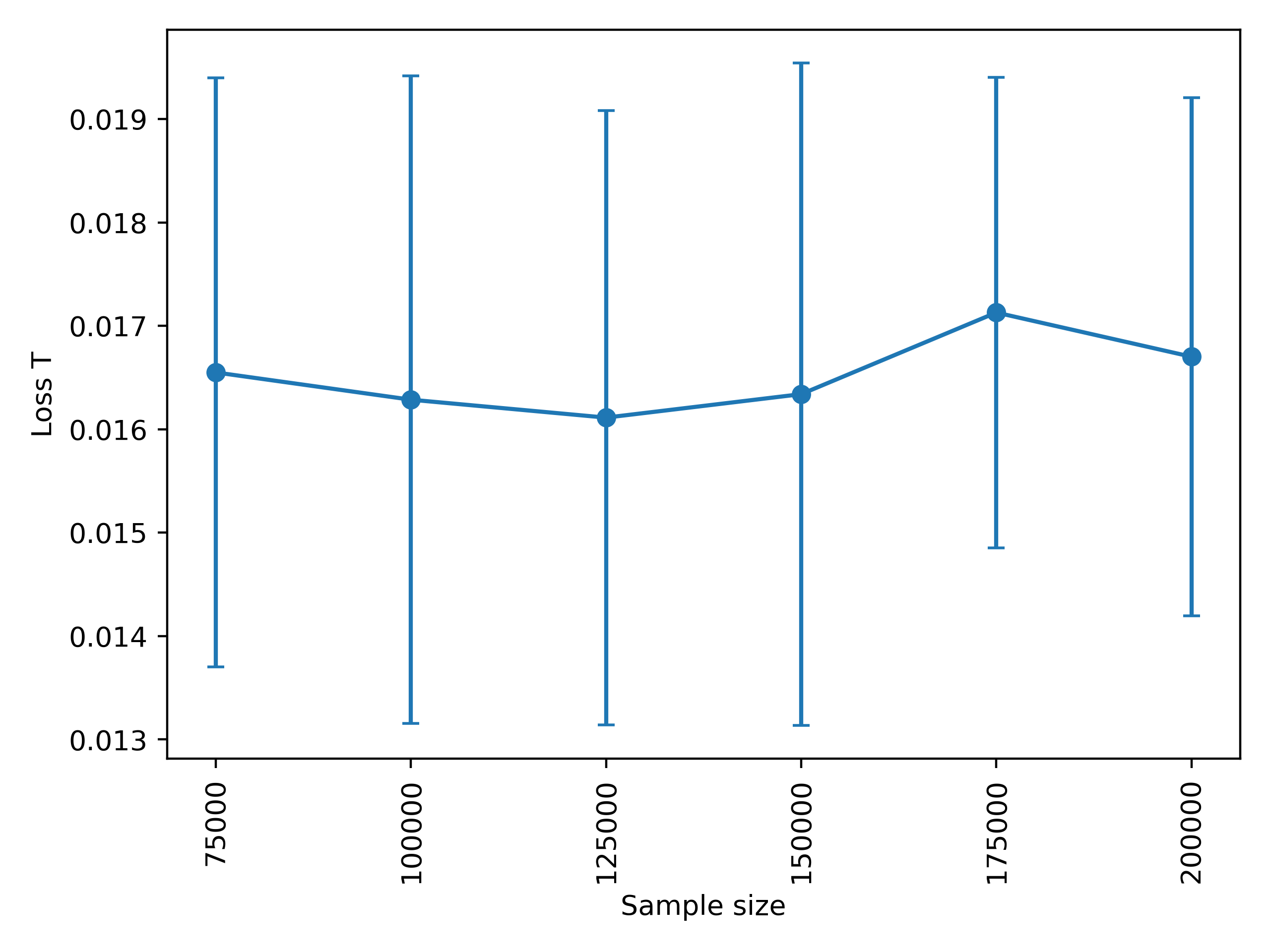"}
    \caption{\label{fig:error_elp_np}\textbf{Loss on $\T$}: The loss on distribution $\T$ until convergence.}
    \ \\
  \end{subfigure}
  \begin{subfigure}[t]{.49\textwidth}
    \centering
    \includegraphics[scale=0.45]{"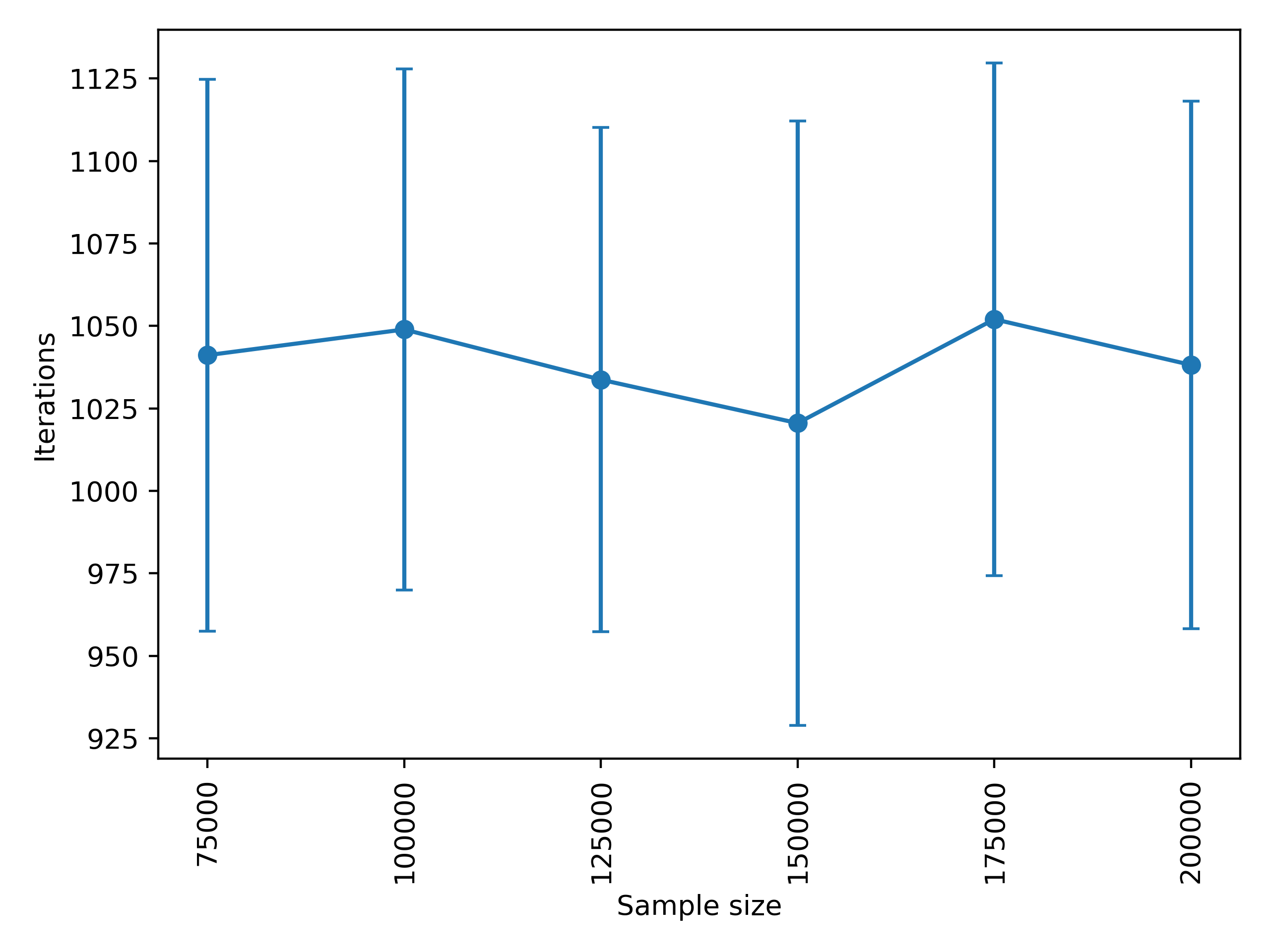"}
    \caption{\label{fig:iterations_elp_np}\textbf{Iterations number}: The iterations number until the convergence.}
    \ \\
  \end{subfigure}
  \medskip
  \begin{subfigure}[t]{.49\textwidth}
    \centering
    \includegraphics[scale=0.5]{"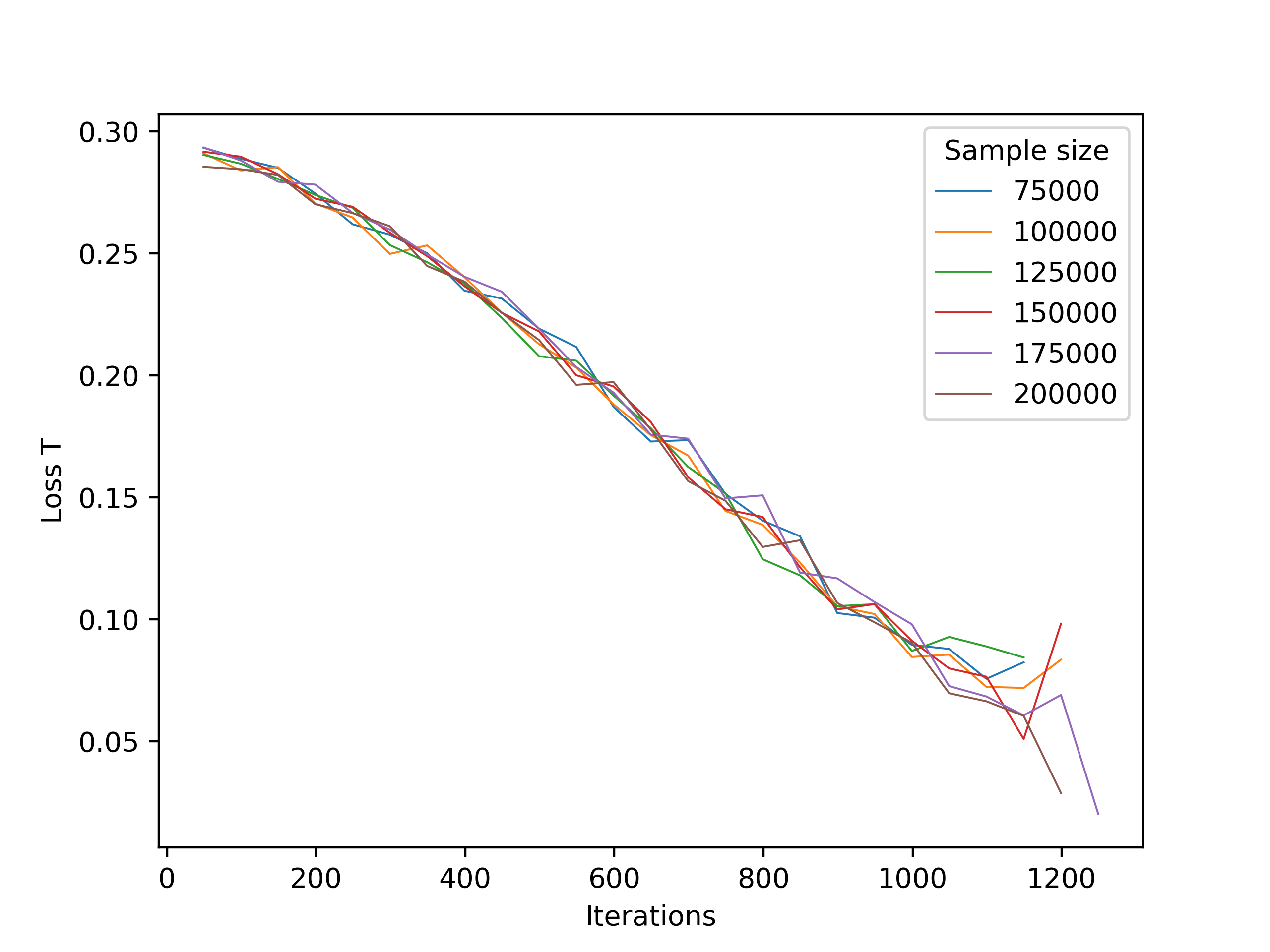"}
    \caption{\label{fig:error_on_run_elp_np}\textbf{Loss on $\T$ along the run}: the loss on $T$ decreases with the iterations.}
  \end{subfigure}
  \caption{\label{fig:all_results_noDP} Empirical Experiment Results}
\end{figure}

Non-privately, we run the online SGD with a maximum of $10^4$ iterations or until finding an exact hyperplane with a loss of at most $\alpha=0.01$ with $\beta=\nicefrac{10^{-6}}{T}$. Aiming to find the sample complexity of the curator-agents for which we get an hypothesis of error $\leq 2\alpha$ over $\T$, we ran our experiments with varying values of $n = \{ 75K, 100K, 125K, 150K, 175K, 200K \}$, repeating each experiment $t=50$ times. Note that all these values of $n$ are below the worst-case bound (in our settings is $15\cdot 10^{12}$), we consistently return a hypothesis of small-loss on $\T$. Results appear in Figure~\ref{fig:error_elp_np}. Again, we see that the number of MW-iterations, $T$, isn't greatly affected by the increasing sample complexity, as seen in Figure~\ref{fig:error_on_run_elp_np}. Also in all values of $n$ we halt after $\approx 1000$ iterations (Results appear in Figure~\ref{fig:iterations_elp_np}) which is far below than the $O(\alpha^{-2})$-upper bound. 


\begin{figure}
\begin{subfigure}[t]{.49\textwidth}
    \centering
    \includegraphics[scale=0.45]{"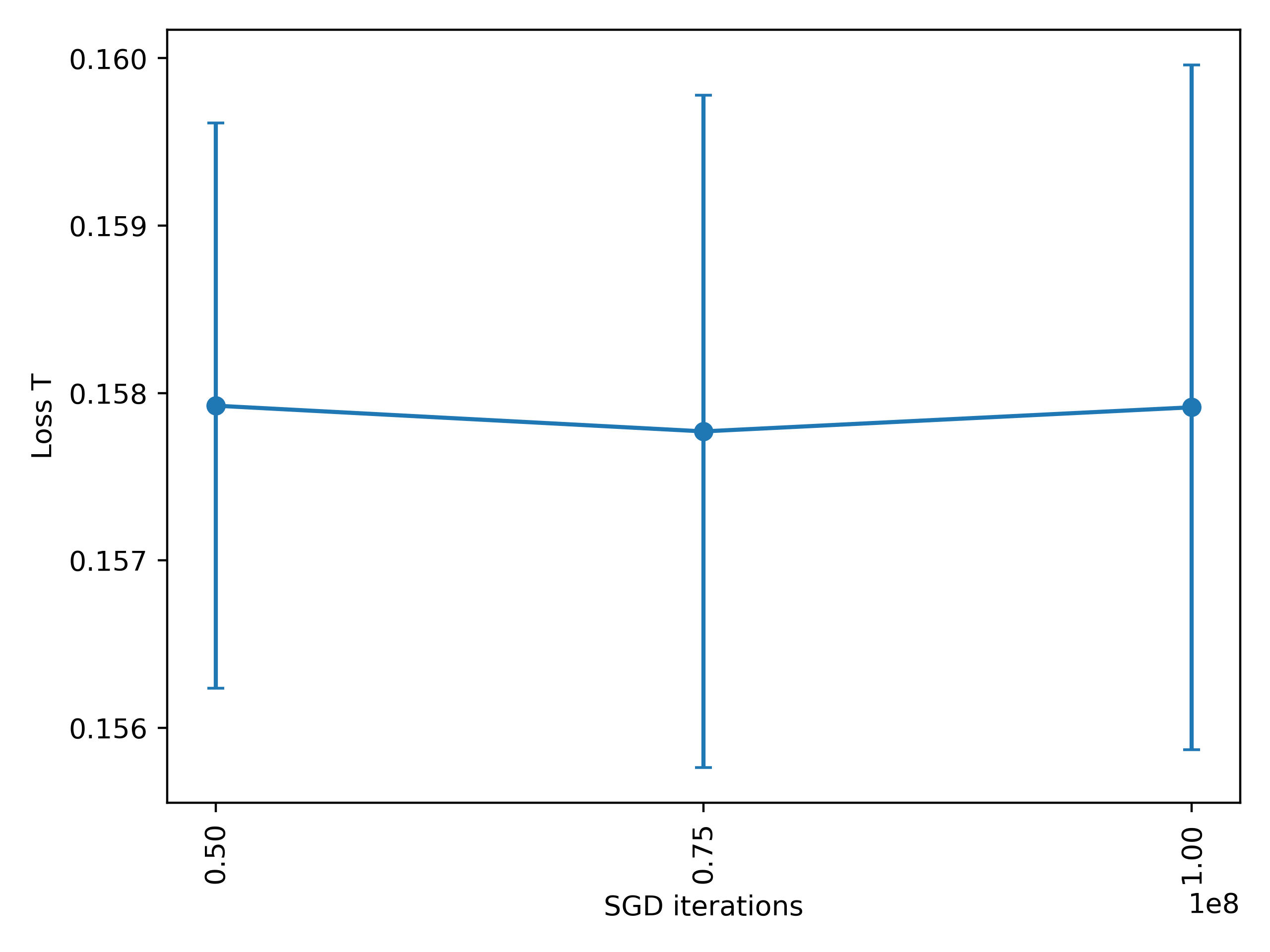"}
    \caption{\label{fig:error_elp_p}\textbf{Loss on $\T$}: The loss on distribution $\T$ until convergence.}
    \ \\
  \end{subfigure}
  \begin{subfigure}[t]{.49\textwidth}
    \centering
    \includegraphics[scale=0.45]{"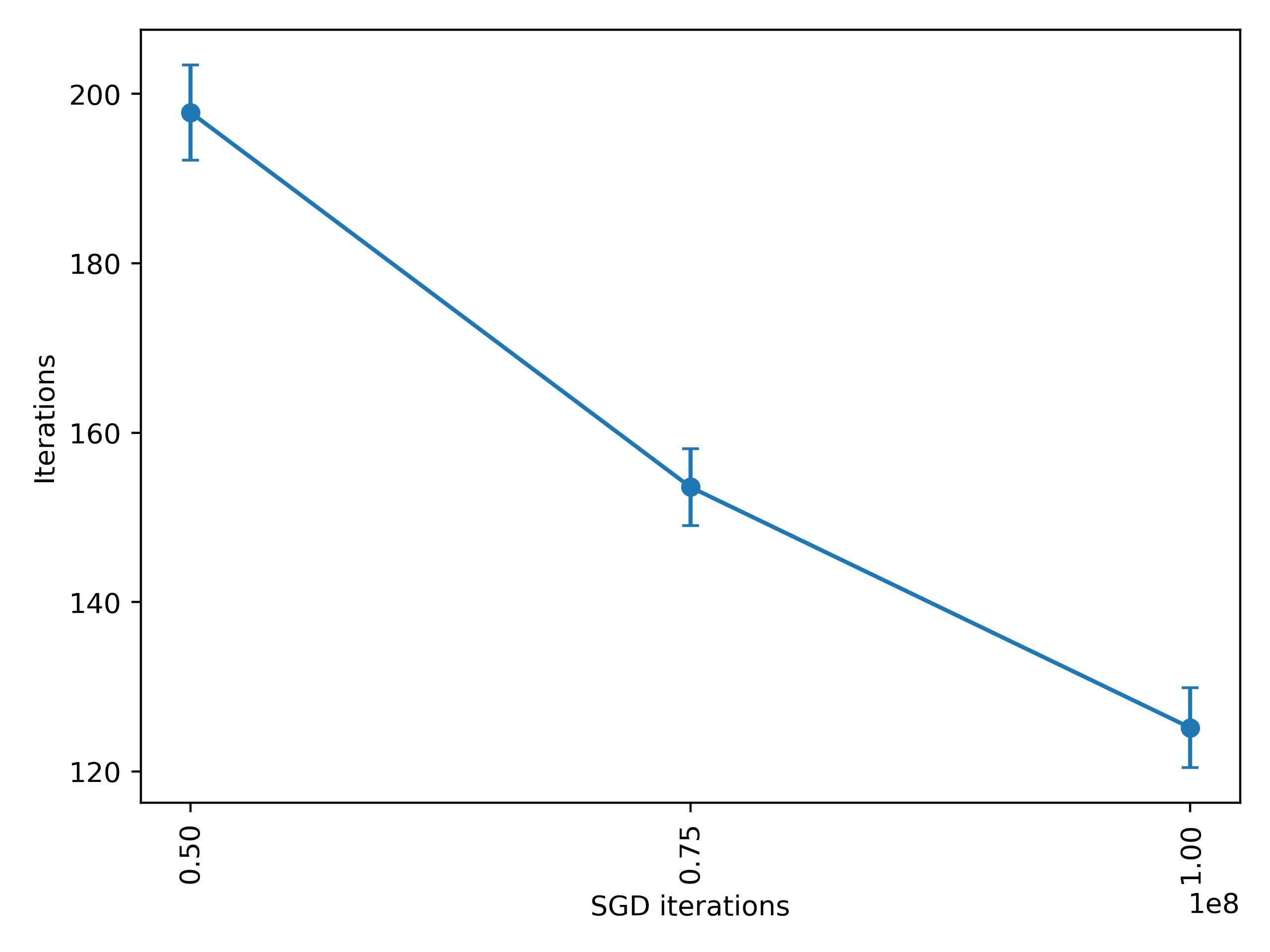"}
    \caption{\label{fig:iterations_elp_p}\textbf{Iterations number}: The MW iterations number until convergence decreases as SGD iterations increases.}
    \ \\
  \end{subfigure}
  \begin{subfigure}[t]{.49\textwidth}
    \centering
    \includegraphics[scale=0.5]{"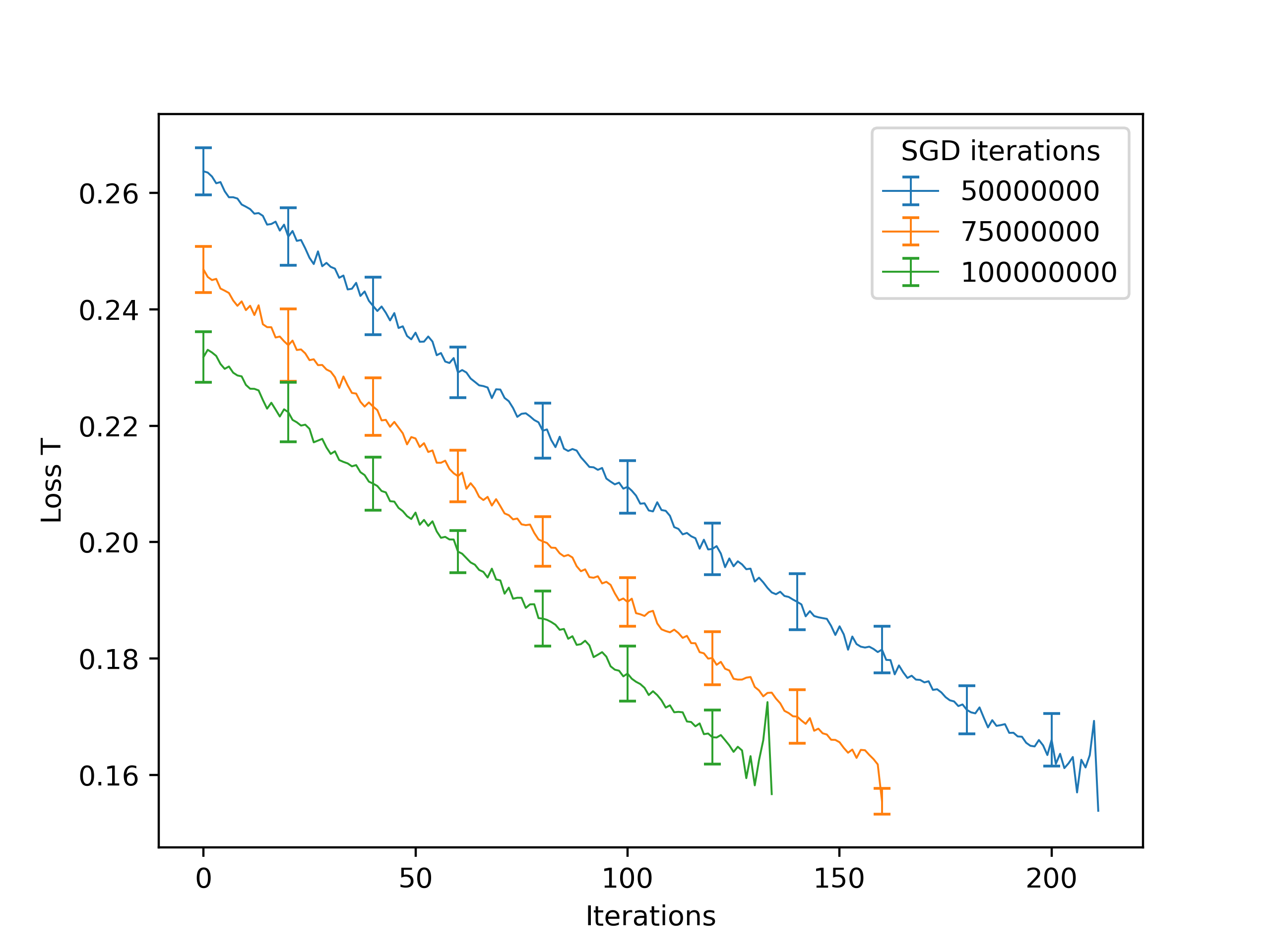"}
    \caption{\label{fig:error_on_run_elp_p}\textbf{Loss on $\T$ along the run}: the loss on $T$ decreases with the iterations.}
  \end{subfigure}
  \caption{\label{fig:all_results_DP} Empirical Experiment Results}
\end{figure}

In the \emph{private} version of the algorithm we applied Algorithm~\ref{alg:private_STR_with_SGD}  with the private online SGD (Algorithm~\ref{alg:private_sgd}). The settings of the non-private version required a vast amount of memory because of the sample complexity bounds that are pretty big even for moderate values of $\alpha$ and $\epsilon$, so we had to change some parameters and use fairly close distributions $\S$ and $\T$. The changes from the non-private settings are as follows. We set the dimension to $d=8$ and the number of ``closer'' coordinates to $k=2$, so  $\S = {\cal N}(\bar 0, (0.1)^2I_d)$ whereas  $\T = {\cal N}(\bar 0, (0.1)^2I_{d-k}\otimes (0.07)^2I_k)$. (It is a matter of simple calculation to show that $\chisq{\T}{\S}+1 = (\frac{ \sigma_S^4 }{\sigma_T^2(2\sigma_S^2-\sigma_T^2)})^{k/2}> 1.35$, where $\sigma_S, \sigma_T$ are the std in the $k$ coordinates of $S$ and $T$ respectively.) We also set $r^0$ is a threshold for $0.4$ of the examples, i.e. $\Pr_{X\sim \chi^2_k}[X < r^0]=0.4$. We set $\alpha=0.08$, and $\kappa=\frac{\alpha}{4(\chi^2+1)}$, and used the privacy parameters of $\epsilon=0.5$ and $\delta=0.0001$. Following similar calculations to before we set the bound on our hypothesis set's diameter as $B=0.006$ and the Lipschitz value of $L=0.1$. 

We succeeded to get from the SGD a hypothesis whose loss is lower than $\alpha=0.08$ with $\beta=\nicefrac{10^{-6}}{T}$ after $50M$ iterations. So we run SGD with $R=50M$, $R=75M$, and $R=100M$ iterations, repeating each experiment $t=50$ time, to see their influence of them on the required number of MW-iterations. Our calculations lead to a sample complexity of  $n=324,700,000$ which we used in all runs. We can see (in Figure~\ref{fig:iterations_elp_p}) that the number of MW-iterations decreases as the iteration of SGD is increasing. In addition, the loss on $\T$ starts higher as the SGD iterations decrease (Figure~\ref{fig:error_on_run_elp_p}). However, in all these runs MW algorithm converged to $2\alpha$ (Figure~\ref{fig:error_elp_p}).

Due to the sample size, we were not able to experiment thoroughly with the private version of our algorithm, alas, our experiments do show that we succeed in implementing the algorithm. Furthermore, we also experimented with a SGD version which minimized the Lasso-regularized version of our algorithm. This version converged in a single iteration --- namely, it found the true separating coordinates on $\S$. This is similar in spirit to the work of~\citet{AventKZHL17} which also used the curator-agents to find the coordinates of the regression.

\newcommand{\cut}[1]{}

\cut{
\newpage

\

\newpage

\begin{figure}
  \begin{subfigure}[t]{.4\textwidth}
    \centering
    \includegraphics[width=\linewidth]{"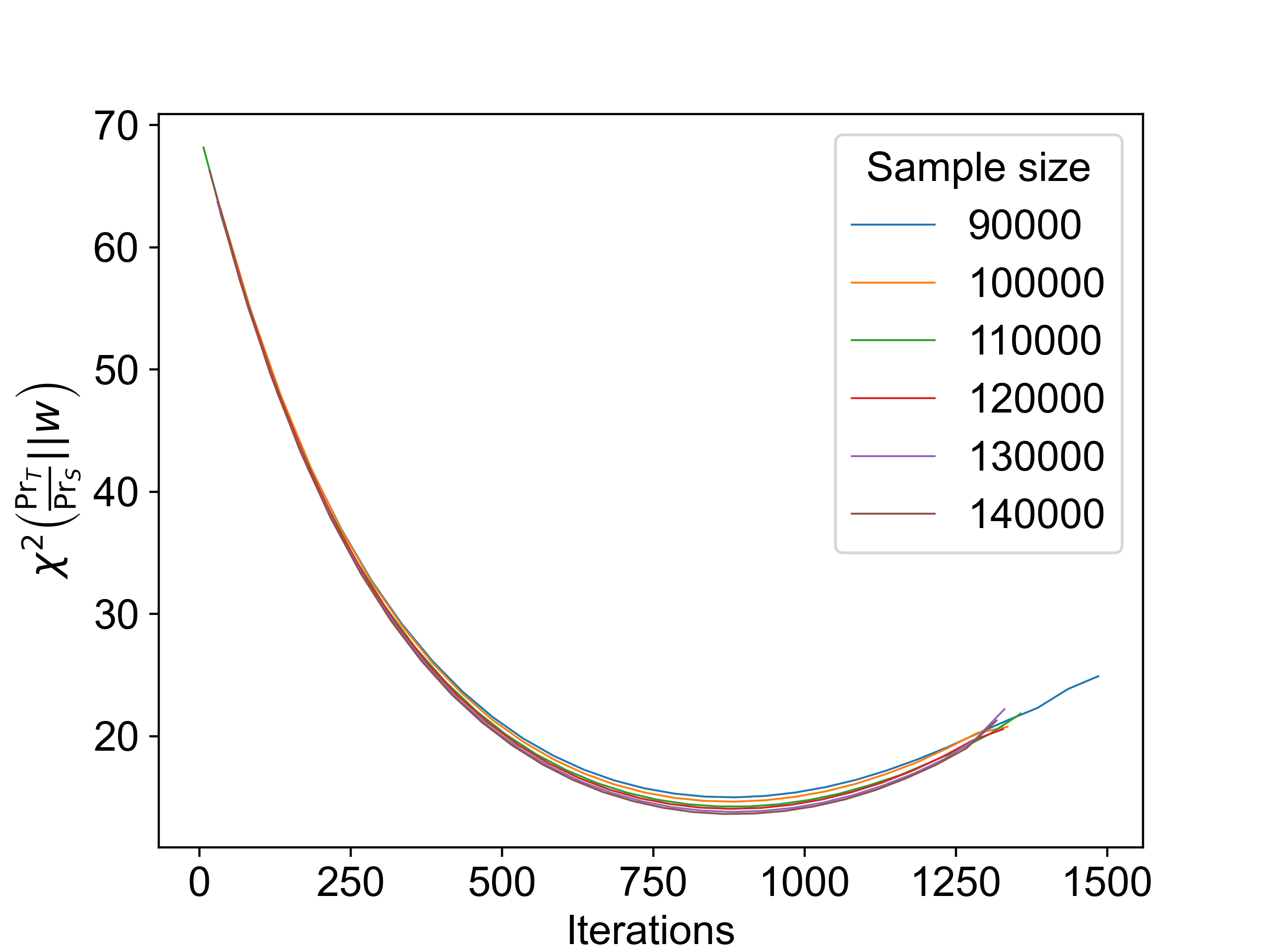"}
    \caption{\textbf{$\chi^2\bigg(\frac{\Pr_T}{\Pr_S}\|w\bigg)$}: the weights that tuning by STR algoritm converges to the importance weights. (\textit{A ohne B}) \osnote{It is strange that in the last iterations chi-seq start to increase (I check in the raw files and it is not programming mistake), there is another figure of TV distance:  \url{TV_distance_vs_iterations_90000-140000.png}}}
  \end{subfigure}
  \hfill
  \end{figure}

\newpage

\section{Refael:}

\subsection{Related Work}

The classic problem of transfer learning has been studied extensively since the 20th century.
One of the first bounds on the error on the $\T$ distribution developed by \citet{ben2010theory} and depends on "symmetric difference hypothesis distance" ($\calH\Delta\calH$-distance), which depend on the best classifiers in a specific hypothesis class and means is the maximum "disagreement" between each pair of hypothesis in the class, where the "disagreement" of pair of hypotheses measured by the difference between the fractions in distribution $\S$ and $\T$ in which the labels of the two hypotheses are different. This distance can be estimated from finite samples from the $\S$ and $\T$ distributions, but appropriate only for binary classification and $L_1$-loss function. \citet{mansour2009domain} developed the discrepancy distance that is extension of $\calH\Delta\calH$-distance also for multi-class classification and regression.

It is worth noting several works that also rely on importance sampling to obtain an hypothesis of small loss w.r.t.~the target distribution. Given the importance weights (or IS) of examples from $\S$, the average of the weighted losses of the hypothesis on the examples from $\S$  (known as $ I_n $) can be estimator of the expectation of the hypothesis loss on $\T$ ( Known as $ I $). IS has been extensively researched in the context of transfer learning and the main problem with this method is the lack of boundary of the weights especially in continuous distributions that do not allow the use of sub-gaussian concentration limits. A good basic example is the word of \citet{chatterjee2017sample} who showed an upper bound of $\exp^{\KL{\T}{\S}}$ on the sample size required for the ability to control the $\E_{x\sim \S}[|I_n-I|]$, and similar lower bound for controlling $\calP(I_n - I \ge \delta), \delta > 0$. Notice that $\exp^{\KL{\T}{\S}} \le \chisq{\T}{\S}+1$ \cite{gibbs2002choosing}, and in some papers \cite{agapiou2017importance} the upper bound pronounced by $\chisq{\T}{\S}+1$. The problem is that $\E_{x\sim \S}[|I_n-I|]$ bounded by the the probability that  $log(\rho(Y))$ (where $\rho(Y)$ is the importance weight of $Y \sim \T$) greater than its expectation $\KL{\T}{\S}$. Admittedly, $log(\rho(Y))$ concentrates around its expectation but the concentration rate is not similar to the rate of sub-gaussian random variables (in which the failure probability ($\beta$) decreases exponentially with the sample size). The seminal result of~\citet{CortesMM10} achieved a bound that is, in spirit, more similar to the desired sub-gaussian behavior. They also proved that with sample size bounds depending on $\chisq{\T}{\S}$ we can achieve accurate estimation of the loss of any hypothesis (simultaneously) in a finite pdim hypothesis class ${\calH}$ over $\T$. Recently, \citet{metelli2021subgaussian} suggest a method of correction of the weights that allow obtaining subgaussian behavior. In the two last methods, prior knowledge of the second moment of the weights is required. Another and more familiar correction is the self-normalization of the weights, which they limited between [0,1] \cite{chatterjee2017sample}, But in this method, the expectation of the average loss in weighted examples from $ S $ is not the expectation of a loss of $\T$, and here also there is no sub-gaussian behavior. A correction we used in our article is to truncate a small portion of the distribution that contains the large weights. The main reason we chose this method is that it is essential to our algorithm for privacy reasons. We use the sample complexity bounds that shown in~\citet{CortesMM10} for such a version of importance weights that assume final $\chisq{\T}{\S}$. and with inequality of \citet{Bernstein54} that analyzed in terms of $\chisq{\T}{\S}+1$.

In order to use the IS method to evaluate the loss on $\T$ we need prior knowledge on the IS or at least a knowledge on the distribution $\S$ and $\T$. In more general cases where access to distribution $\T$ is restricted to unlabeled examples, there are several methods for evaluating IS summarized in \citet{maia2022effective}. But in our case, we have no access to the examples of distribution $\T$, but only to \emph{${\sf SQ}_{\tau}$-oracle} which returns the noisy loss of a given hypothesis, and we need evaluate IS according to these answers. For this task, we use the Multiplicative-Weights algorithm. In fact, we do not really need to achieve the importance weights, but we use the fact that they exist to ensure the stopping conditions of the algorithm, while the real convergence may be at another point.

One of the earlier works that used a boosting algorithm (based on the MW algorithm) for transfer learning, is that of \citet{5539857}. In their work, the distributions $\S$ and $\T$ are identical but the features of the classes are different. In each iteration, the learning is made on a main sample from $\T$ combined with an auxiliary sample from $\S$, and the tuning of the weights of the two samples is made using a variation of boosting algorithm that focuses the algorithm on the example from $\T$ and on the correct predicted examples from $\S$. More recent work is "gapBoost" algorithm that also gets samples from $\S$ and $\T$ and uses the boosting algorithm for tuning the weights. Clearly, our work is substantially different since we do not use such assumptions on $\S$ and $\T$ and we have access to $\T$ distribution only by {\sf SQ} model.

In fact, the basic idea to use the MW algorithm for transfer learning stems from the proof (one of many proofs) of the algorithm. The proof was given by~\citet{AroraHK12} which gives a good insight into the course of action of the algorithm, is based on the (KL)-divergence. The proof argues that in each iteration, the distribution of the examples gets closer (in sense of KL-divergence) to some target distribution. In this way, we can bring the weights closer to IS and derive from this measure the expectation of the average loss of the weighted examples.

However, to the best of our knowledge, no previous work has studied the unique $\S$-through-{\sf PAC} and $\T$-through-{\sf SQ} model we present in Section~\ref{sec:transfer_learning_non_private}.

\subsection{Inequalities}
\begin{definition}
\label{def:w}
Let $w(x):\calX\to \R_+$ be the function defined by $w(x) = \frac{\pdfT(x)}{\pdfS(x)}$, where $\pdfT$ and $\pdfS$ denotes the PDF of $\T$ and $\S$ respectively. We call $w(x)$ the \emph{weight of $x$}.
\end{definition}

\begin{definition}
\label{def:u}
Lets $u(x) = \min (w(x), \frac{4(\chi^2+1)}{\alpha})$ truncated weights of $x$ and $u^\prime(x) = \frac{u(x)}{\nicefrac{4(\chi^2+1)}{\alpha}}$ truncated weights of $x$ upper bounded by 1.
\end{definition}

\begin{lemma}
\label{lem:equations}
The following equations hold:

\begin{equation}
\label{eq:ETw}
    \ET[w(x)] = \int_{\calX} \pdfT(x) \frac{\pdfT(x)}{\pdfS(x)} \, dx  = \chi^2 + 1
\end{equation}

\begin{equation}
\label{eq:ESw}
    \ES[w(x)] = \int_{\calX} \pdfS(x) \frac{\pdfT(x)}{\pdfS(x)} \, dx = \int_{\calX} \pdfT(x) \, dx = 1
\end{equation}

\begin{equation}
\label{eq:ESw2}
    \ES[w(x)^2] = \int_{\calX} \pdfS(x) \bigg(\frac{\pdfT(x)}{\pdfS(x)}\bigg)^2 \, dx = \chi^2 + 1
\end{equation}

\begin{equation}
\label{eq:ESu2}
    \ES[u(x)^2] \le \ES[w(x)^2] = \chi^2 + 1
\end{equation}

\begin{equation}
\label{eq:ESu}
\begin{aligned}
    1 &= \ES[w(x)] \ge \ES[u(x)] \ge \ES[u(x)\1_{\A^c}] \\
    &= \ES[w(x)\1_{\A^c}] = \ES[w(x)]-\ES[w(x)\1_{\A}] \\
    &\ge 1 -\nicefrac{\alpha}{4}
\end{aligned}
\end{equation}

Let $\A = \{x; w(x) \ge \frac{4(\chi^2 +1)}{\alpha}\}$:

\begin{equation}
\label{eq:ESwA}
\begin{aligned}
  \ES[w(x)\1_{\A}] &= \int_{\calX} \pdfT(x)\1_{\A} \, dx \\
  &= \calP_T \bigg(w(x) \ge \frac{4(\chi^2+1)}{\alpha} \bigg) \\ 
  &\le \frac{\alpha \E\nolimits_{x\sim\T}[w(x)]}{4(\chi^2+1)} = \nicefrac{\alpha}{4}
\end{aligned}
\end{equation}

\begin{equation}
\label{eq:ESw-u}
\begin{aligned}
    &\ES[\big|w(x)-u(x)\big|\Loss(h(x),\ell(x))] \\
    &\le \ES[|w(x)-u(x)|] \le \ES[w(x)] \\
    &\le \ES[w(x)\1_{\A}] \le \nicefrac{\alpha}{4}
\end{aligned}
\end{equation}

Let $r(x) = 1-\Loss(h(x),\ell(x))$:

\begin{equation}
\label{eq:ESur}
\begin{aligned}
    \ES[u(x)r(x)] &\le \ES[u(x)r(x)\1_{\A^c}] + \ES[u(x)r(x)\1_{\A}] \\
    &=  \ES[w(x)r(x)\1_{\A^c}] + \ES[u(x)\1_{\A}] \\
    &\le \ES[w(x)r(x)\1_{\A^c}] + \ES[w(x)\1_{\A}] \\
    &\le \ET[r(x)\1_{\A^c}] + \nicefrac{\alpha}{4} \\
    &\le \ET[r(x)] + \nicefrac{\alpha}{4}
\end{aligned}
\end{equation}
\end{lemma}

\begin{lemma}
\label{lem:MEANur-ESur}
For each iteration of the algorithm and for each hypothesis of $\calH$, given $n = O\bigg(\frac{(\chi^2+1)\ln(\nicefrac{|\calH|}{\beta})}{\alpha^2}\bigg)$, w.p. at least $1-\nicefrac{\beta}{2}$:
\begin{equation}
    \frac{1}{n} \sum_{i=1}^n u(x_i)r(x_i) - \ES[u(x)r(x)] \le \frac{\alpha}{8}
\end{equation}
\end{lemma}

Proof:

Using the Bernstein Inequality (\ref{bernstein-ieq}), with $n$ i.i.d. zero-mean random variables $u(X)r(X) - \ES[u(X)r(X)]$ comes from $\{x_i\}_{i=1}^n$ i.i.d. examples from distribution $\S$. Then, the upper bound is \[d = \frac{4(\chi^2 + 1)}{\alpha} - \ES[u(X)r(X)]\] and the second moment is:
\begin{equation*}
    \begin{aligned}
        \ES\big[\big(&u(X)r(X)-\ES[u(X)r(X)]\big)^2\big] \\
        &= \ES[u(X)^2r(X)^2]- \ES[u(X)r(X)]^2 \\
        &\le \ES[u(X)^2r(X)^2] \le \ES[u(X)^2] \\
        & \underset{\substack{\text{\cref{eq:ESu2}}}}{\le} \chi^2+1
    \end{aligned}
\end{equation*}

Set in Bernstein's inequality:

\begin{equation*}
\begin{aligned}
    &\calP \bigg(\frac{1}{n}\sum_n u(x_i)r(x_i) - \ES[u(X)r(X)] \ge \frac{\alpha}{8} \bigg) =\\
    &\calP \bigg(\sum_n u(x_i)r(x_i) - \ES\big[\sum_n u(x_i)r(x_i)\big] \ge \frac{\alpha n}{8} \bigg) \le \\
    & \exp \bigg(\frac{-\nicefrac{\alpha^2n^2}{64}}{2n(\chi^2+1) + \frac{2}{3}(\frac{\alpha n}{8})\big(\frac{4(\chi^2 + 1)}{\alpha} - \ES[u(X)r(X)]\big)}\bigg) \\
    & \le \exp \bigg(\frac{-\nicefrac{\alpha^2n^2}{64}}{2n(\chi^2+1) + \frac{ n(\chi^2 + 1)}{3}}\bigg) = \exp \bigg(\frac{-3\alpha^2n}{448(\chi^2+1)}\bigg)
\end{aligned}
\end{equation*}

The inequality holds for the all iterations simultaneously because the algorithm uses the same sample in all iterations. We require the probability $\le \nicefrac{\beta}{2|\calH|}$ and then using union bound the total risk probability over all hypothesis in $\calH$ is $\nicefrac{\beta}{2}$. Rearranging we obtain the bound:
\begin{equation*}
    n \ge \frac{448(\chi^2+1)\ln(\nicefrac{2|\calH|}{\beta})}{3\alpha^2}
\end{equation*}

\begin{lemma}
\label{lem:SuLESS2n}
Given $n = O\bigg((\chi^2+1)\ln(\nicefrac{1}{\beta^\prime})\bigg)$, w.p. at least $1-\beta^\prime$:
\begin{equation}
    \sum_{i=1}^n u(x_i) \le 2n 
\end{equation}
\end{lemma}

Proof:

In similar way to the proof of \cref{lem:MEANur-ESur}, Using the Bernstein Inequality (\ref{bernstein-ieq}), with $n$ i.i.d. zero-mean random variables $u(X) - \ES[u(X)]$ comes from $\{x_i\}_{i=1}^n$ i.i.d. examples from distribution $\S$. Then, the upper bound is \[d = \frac{4(\chi^2 + 1)}{\alpha} - \ES[u(X)]\] and the second moment is:
\begin{equation*}
    \begin{aligned}
        \ES\big[\big(&u(X)-\ES[u(X)]\big)^2\big] \\
        &= \ES[u(X)^2]- \ES[u(X)]^2 \\
        &\le \ES[u(X)^2] \underset{\substack{\text{\cref{eq:ESu2}}}}{\le} \chi^2+1
    \end{aligned}
\end{equation*}

Set in Bernstein's inequality:

\begin{equation*}
\begin{aligned}
    &\calP \bigg(\sum_n u(x_i) - n\ES\big[u(X)\big] \ge n \bigg) \le \\
    & \exp \bigg(\frac{-n^2}{2n(\chi^2+1) + \frac{2}{3}(\frac{\alpha n}{8})\big(\frac{4(\chi^2 + 1)}{\alpha} - \ES[u(X)]\big)}\bigg) \\
    & \le \exp \bigg(\frac{-n^2}{2n(\chi^2+1) + \frac{ n(\chi^2 + 1)}{3}}\bigg) = \exp \bigg(\frac{-3n}{448(\chi^2+1)}\bigg)
\end{aligned}
\end{equation*}

We require the probability $\le \beta^\prime$ and rearranging we obtain the bound:
\begin{equation}
\label{lem:SuLESS2n-exact}
    n \ge \frac{448(\chi^2+1)\ln(\nicefrac{1}{\beta^\prime})}{3}
\end{equation}

Plug-in the upper bound of 1 on $\ES[u(X)]$ \cref{eq:ESu} and get:

\begin{equation*}
\begin{aligned}
    &\calP \bigg(\sum_n u(x_i) \ge 2n \bigg) \le \\
    &\calP \bigg(\sum_n u(x_i) - n \ge n \bigg) \underset{\substack{\text{\cref{eq:ESu}}}}{\le} \\
    &\calP \bigg(\sum_n u(x_i) - n\ES\big[u(X)\big] \ge n \bigg) \le \\
    &1-\beta^\prime
    \end{aligned}
\end{equation*}

\begin{lemma}[lemma 4 in \citep{CSABA}]
\label{lem:Csaba}
Let $\{X_I\}_{i=1}^n$  be i.i.d. non-negative random variables (not necessarily bounded from above). Then, for any $t \in \big[0, n\E[X]\big]$ with probability at least $1-\beta^\prime$,
\begin{equation}
    \sum_{i=1}^n X_i \ge n\E[X] - \sqrt{2n\ln(\nicefrac{1}{\beta^\prime})\E[X^2}]
\end{equation}
\end{lemma}

The proof is given in \citep{CSABA} in appendix C.

\begin{lemma}
\label{lem:NORMur-MEANur}
For each iteration of the algorithm and for each hypothesis of $\calH$, given $n = O\bigg(\frac{(\chi^2+1)\ln(\nicefrac{|\calH|}{\beta})}{\alpha^2}\bigg)$, w.p. at least $1-\nicefrac{\beta}{2}$:
\begin{equation}
    \frac{1}{\sum_{i=1}^n u(x_i)} \sum_{i=1}^n u(x_i)r(x_i)-\frac{1}{n} \sum_{i=1}^n u(x_i)r(x_i) \le \alpha
\end{equation}
\end{lemma}

Proof:

Using \cref{lem:SuLESS2n} with setting $\beta^\prime = \nicefrac{\beta}{4|\calH|}$, given $n \ge \frac{448(\chi^2+1)\ln(\nicefrac{4|\calH|}{\beta})}{3}$, with the union bound on the all hypothesis in $\calH$, w.p. at least $1-\nicefrac{\beta}{4}$, 

\begin{equation}
\label{eq:lemProofPart1-NORMur-MEANur}
    \begin{aligned}
        &\frac{1}{\sum_{i=1}^n u(x_i)} \sum_{i=1}^n u(x_i)r(x_i)-\frac{1}{n} \sum_{i=1}^n u(x_i)r(x_i) = \\
        &\frac{1}{n} \sum_{i=1}^n u(x_i)r(x_i)\bigg(\frac{n}{\sum_{i=1}^n u(x_i)} -1 \bigg) \le \\
        &\frac{1}{n} \sum_{i=1}^n u(x_i)\bigg(\frac{n}{\sum_{i=1}^n u(x_i)} - 1 \bigg) \underset{\substack{\text{\cref{lem:SuLESS2n}}}}{\le} \\
        &2\bigg(\frac{n}{\sum_{i=1}^n u(x_i)} -1 \bigg)
    \end{aligned}
\end{equation}

We require that $2\bigg(\frac{n}{\sum_{i=1}^n u(x_i)} -1 \bigg) \le \alpha$, i.e. $\sum_{i=1}^n u(x_i) \ge (\frac{2}{\alpha+2})n$ w.p. at least $1-\nicefrac{\beta}{4}$.

\begin{remark}
It is impossible to bound $2\bigg(\frac{n}{\sum_{i=1}^n u(x_i)} -1 \bigg)$ with lower value than $\alpha$, because a collision with the following condition. 
\end{remark}

The last requirement can be achieved using \cref{lem:Csaba} with setting $\beta^\prime = \nicefrac{\beta}{4}$, w.p. at least $1-\nicefrac{\beta}{4}$,
\[
    \sum_i^n u(x_i) \ge n\ES[u(x)] - \sqrt{2n\ln(\nicefrac{4}{\beta})\ES[u(x)^2)]}
\]

Plug-in the bounds on $\ES[u(x)]$ (\cref{eq:ESu}) and $\ES[u(x)^2]$ (\cref{eq:ESu2}) and get w.p. at least $1-\nicefrac{\beta}{4}$,

\[
    \sum_i^n u(x_i) \ge n(1-\nicefrac{\alpha}{4}) - \sqrt{2n\ln(\nicefrac{4}{\beta})(\chi^2+1)}
\]

We can conclude that w.p. at least $1-\nicefrac{\beta}{4}$,

\begin{equation}
\label{eq:lemProofPart2-NORMur-MEANur}
    \sum_{i=1}^n u(x_i) \ge (\frac{2}{\alpha+2})n
\end{equation}

conditioned on the fact that:

\begin{equation*}
    \begin{aligned}
        &(\frac{2}{\alpha+2})n \le (1-\nicefrac{\alpha}{4})n - \sqrt{2n\ln(\nicefrac{4}{\beta})(\chi^2+1)} \\
        &\Rightarrow \sqrt{2n\ln(\nicefrac{4}{\beta})(\chi^2+1)} \le  \frac{\alpha(2-\alpha)}{4(2+\alpha)}n \le  \frac{\alpha}{4} \\
        &\Rightarrow n \ge \frac{32\ln(\nicefrac{4}{\beta})(\chi^2+1)}{\alpha^2}
    \end{aligned}
\end{equation*}

Use the union bound on \cref{eq:lemProofPart1-NORMur-MEANur} and \cref{eq:lemProofPart2-NORMur-MEANur} to complete the proof.

\begin{theorem}
\label{lem:NORMur-ETr}
For each iteration of the algorithm and for each hypothesis of $\calH$, given $n = O\bigg(\frac{(\chi^2+1)\ln(\nicefrac{|\calH|}{\beta})}{\alpha^2}\bigg)$, w.p. at least $1-\beta$:
\begin{equation}
    \frac{1}{\sum_{i=1}^n u(x_i)} \sum_{i=1}^n u(x_i)r(x_i)-\ET[r(X)] \le 2\alpha
\end{equation}
\end{theorem}

Proof:

Use the triangle inequality between:

\begin{enumerate}
\item \cref{eq:ESur}: 

\[
\ES[u(x)r(x)] - \ET[r(x)] \le \nicefrac{\alpha}{4}
\]

\item \cref{lem:MEANur-ESur}: given $n = O\bigg(\frac{(\chi^2+1)\ln(\nicefrac{|\calH|}{\beta})}{\alpha^2}\bigg)$, w.p. at least $1-\nicefrac{\beta}{2}$:

\[
\frac{1}{n} \sum_{i=1}^n u(x_i)r(x_i) - \ES[u(x)r(x)] \le \frac{\alpha}{8}
\]

\item \cref{lem:NORMur-MEANur}: given $n = O\bigg(\frac{(\chi^2+1)\ln(\nicefrac{|\calH|}{\beta})}{\alpha^2}\bigg)$, w.p. at least $1-\nicefrac{\beta}{2}$

\[
\frac{1}{\sum_{i=1}^n u(x_i)} \sum_{i=1}^n u(x_i)r(x_i)-\frac{1}{n} \sum_{i=1}^n u(x_i)r(x_i) \le \alpha
\]
\end{enumerate}

To get that given $n = O\bigg(\frac{(\chi^2+1)\ln(\nicefrac{|\calH|}{\beta})}{\alpha^2}\bigg)$  w.p. at least $1-\beta$:

\[
\frac{1}{\sum_{i=1}^n u(x_i)} \sum_{i=1}^n u(x_i)r(x_i) - \ET[r(x)] \le \frac{11\alpha}{8} \le 2\alpha
\]

\begin{lemma}
\label{lem:Su_primeGEkn}
Let $u$ and $u^\prime$ set of measures on $n$ examples as defined in \cref{def:u} and let $\kappa \le 1$, given $n = O\bigg(\frac{(\chi^2+1) \ln(\nicefrac{1}{\beta^\prime)}}{\alpha^2}\bigg)$, w.p. at least $1-\beta^\prime$,
\begin{equation*}
\sum_{i=1}^n u^\prime(x_i) = \frac{\sum_{i=1}^n u(x_i)}{\nicefrac{4(\chi^2+1)}{\alpha}} \ge \kappa n
\end{equation*}
\end{lemma}

Proof:

Using \cref{lem:Csaba}; w.p. at least $1-\beta^\prime$,
\[
    \sum_i^n u(x_i) \ge n\ES[u(x)] - \sqrt{2n\ln(\nicefrac{1}{\beta^\prime})\ES[u(x)^2)]}
\]

Plug-in the bounds on $\ES[u(x)]$ (\cref{eq:ESu}) and $\ES[u(x)^2]$ (\cref{eq:ESu2}) and get w.p. at least $1-\beta^\prime$,

\[
    \sum_i^n u(x_i) \ge n(1-\nicefrac{\alpha}{4}) - \sqrt{2n\ln(\nicefrac{1}{\beta^\prime})(\chi^2+1)}
\]

We can conclude that w.p. at least $1-\beta^\prime$,

\begin{equation}
\label{eq:lemProofPart2-NORMur-MEANur}
    \sum_{i=1}^n u(x_i) \ge \frac{4\kappa n (\chi^2+1)}{\alpha}
\end{equation}

conditioned on the fact that:

\begin{align}
        &\frac{4\kappa n (\chi^2+1)}{\alpha} \le n\left[(1-\nicefrac{\alpha}{4}) - \sqrt{\frac{2\ln(\nicefrac{1}{\beta^\prime})(\chi^2+1)}n}\right] \cr
        \intertext{Since our bound on $n$ gives that}
        &\sqrt{\frac{2\ln(\nicefrac{1}{\beta^\prime})(\chi^2+1)}n} < \frac \alpha 4\cr
        \intertext{then it suffices to set}
        &\kappa \le \frac{\alpha }{8(\chi^2+1)} \le \frac{\alpha (1-\nicefrac{\alpha}{2} )}{4(\chi^2+1)}
\end{align}

\begin{lemma}
\label{lem:kl-u-k}
Let $u$ and $u^\prime$ set of measures on $n$ examples as defined in \cref{def:u}, and $\bar{\kappa}$ another set of uniform measures of size $\kappa \le 1$. 
Then, given $n = O\bigg(\frac{(\chi^2+1) \ln(\nicefrac{1}{\beta^\prime)}}{\alpha^2}\bigg)$, w.p. at least $1-\beta^\prime$,

\begin{equation}
    \KL{u^\prime(X)}{\bar{\kappa}} \le \frac{2 \alpha n \, \log_2(\nicefrac{1}{\kappa})}{4(\chi^2+1)}
\end{equation}
\end{lemma}

Proof:

\begin{equation}
\label{lem:proofPart1-kl-u-k}
\begin{aligned}
    &\KL{u^\prime(X)}{\bar{\kappa}} \underset{\substack{\text{\cref{kl-measures}}}}{=} \\
    &\sum_{i=1}^n u^\prime(x_i) \log_2 \bigg( \frac{u^\prime(x_i)}{\kappa}\bigg) - \sum_{i=1}^n u^\prime(x_i) + \kappa n 
\end{aligned}
\end{equation}

Using \cref{lem:Su_primeGEkn}, given $n = O\bigg(\frac{(\chi^2+1) \ln(\nicefrac{1}{\beta^\prime)}}{\alpha^2}\bigg)$, w.p. at least $1-\beta^\prime$, \, 
$\sum_{i=1}^n u^\prime(x_i) \ge \kappa n$

Therefore, 

\begin{equation}
\label{lem:proofPart2-kl-u-k}
\begin{aligned}
    \cref{lem:proofPart1-kl-u-k} &\le \sum_{i=1}^n u^\prime(x_i) \log_2 \bigg(\frac{u^\prime(x_i)}{\kappa}\bigg) \\
    &\underset{\substack{u^\prime(x) \le 1}}{\le} 
    \sum_{i=1}^n u^\prime(x_i) \log_2 (\nicefrac{1}{\kappa}) \\
    &\underset{\substack{u^\prime(x) \le 1}}{=} 
    \frac{\alpha \log_2(\nicefrac{1}{\kappa})}{4(\chi^2 +1)}\sum_{i=1}^n u(x_i) 
\end{aligned}
\end{equation}

Using \cref{lem:SuLESS2n-exact}, setting $\beta^\prime = \nicefrac{\beta}{2}$, given $n \ge \frac{448(\chi^2+1)\ln(\nicefrac{2}{\beta})}{3}$, w.p. at least $1-\nicefrac{\beta}{2}$, \, $\sum_{i=1}^n u(x_i) \le 2n$ we get:

\begin{equation}
\label{lem:proofPart3-kl-u-k}
\begin{aligned}
    \cref{lem:proofPart2-kl-u-k} &\le \frac{2 \alpha n \log_2(\nicefrac{1}{\kappa})}{4(\chi^2 +1)} 
\end{aligned}
\end{equation}

\section{Private Stochastic Gradient Descent}
\label{sec:refael_privacy_sgd}

Here we use the general online algorithm for agnostic learning from \citep{Hazan16} (algorithm 26) for learning with Stochastic Gradient Descent (SGD). We use in theorem 3.1 with the update rule of algorithm 7 in \citep{Hazan16}.

\begin{algorithm}
   \caption{Online SGD}
   \label{alg:online_sgd}
\begin{algorithmic}
   \STATE {\bfseries Input:} parameters $\alpha^\prime$,$\beta^\prime > 0$, $\sigma \ge \sqrt{20}L$, $\eta_t$, $T$. Under the settings in \cref{lem:sgd-alg}.
   \STATE Let $h_1 \gets$ any $h \in \calH$.  
   \STATE Set $T\gets O\bigg(\max\bigg(\frac{36(L^2+\sigma^2\sqrt{d})}{\alpha^{\prime2}}, \frac{32 \ln(\nicefrac{2}{\beta^\prime})}{\alpha^{\prime2}}\bigg)\bigg)$
   \FOR {($t = 1,2,3,...,T$)}
   \STATE Draw labeled example $(x_t, y_t) \sim \mu$
   \STATE Let $\tilde{\nabla}_t = M_G\big(f_{(x_t, y_t)}(h_t)\big)$
   \STATE $h_{t+1} = \Pi_{\calH} (h_t -\eta_t \tilde{\nabla}_t)$
   \ENDFOR
   \STATE
   \textbf{Return} $\bar{h} = \frac{1}{T} \sum_{t=1}^T h_t$.
\end{algorithmic}
\end{algorithm}

\begin{lemma}
\label{lem:sgd-alg}
Given: distribution $\mu$ on $n$ labeled examples $(x_1, y_1), ..., (x_n, y_n)$ where $y_i$ is the true label $\ell(x_i)$, convex class $\calH \subseteq \R^d$ of hypotheses $h$ such that $\forall \, (h_i, h_j) \in \calH$, $\| h_i - h_j \| \le D$, $L$-lipschitz and convex loss function $f_{(x, y)}(h) = \Loss(h(x), y)$ such that $\err(h^*) = \E\limits_{x\sim\mu}[\Loss(h^*(x),\ell(x))] \le \frac{\alpha^\prime}{2}$ and $\textit{regret}_T(f) = \sum_{t=1}^T \big(f_{(x_t, y_t)}(h_t) - f_{(x_t, y_t)}(h^*)\big)$, a Gaussian mechanism $M_G: f_{(x, y)}(h) \to \nabla f_{(x, y)}(h) + \calN(0, \sigma^2)$, $\tilde{\nabla} = M_G\big(f_{(x, y)}(h)\big)$ such that $\E[\|\tilde{\nabla}\|^2] \le L^2 + d\sigma^2$, and the convex function $f_t^\prime(h) = \langle \tilde{\nabla}_t, h\rangle$, then, with parameter $\eta_t=\frac{D}{\sigma\sqrt{t}}$ w.p. at least $\nicefrac{1}{\beta^\prime}$, after $T \ge \max\bigg(\frac{L^2+\sigma^2\sqrt{d}}{\alpha^{\prime2}}, \frac{\ln(\nicefrac{2}{\beta^\prime})}{\alpha^{\prime2}}\bigg)$ iterations of \cref{alg:online_sgd}, $\err(\bar{h}) \le \alpha^\prime$.
\end{lemma}

Proof:

In \citep{Hazan16} proved that under the given settings, w.p. at least $\nicefrac{1}{\beta^\prime}$, 

\begin{equation}
\label{eq:regret_f_general}
    \err(\bar{h}) \le \err(h^*) + \frac{\textit{regret}_T(f)}{T} + \sqrt{\frac{8 \ln(\nicefrac{2}{\beta^\prime})}{T}}
\end{equation}

According to theorem 3.1 in \citep{Hazan16}, under the above settings, $\textit{regret}_T(f^\prime) = \sum_{t=1}^T \big(f^\prime_t(h_t) - f^\prime_t(h^*)\big) \le \frac{3}{2}D\sqrt{(L^2+d\sigma^2)T}$. Therefore, 

\begin{align}
\label{eq:regret_f_prime}
    \textit{regret}_T(f) &= \sum_{t=1}^T \big(f_{(x_t, y_t)}(h_t) - f_{(x_t, y_t)}(h^*)\big) \cr
    &\le 
    \sum_{t=1}^T\bigg(\big\langle \nabla f_{(x_t, y_t)}(h_t), h_t - h^* \big\rangle \bigg) \tag{convexity of $f$} \\
    &= \sum_{t=1}^T\bigg(\big\langle \E_{\tilde{\nabla}_t\sim\calN\big(\nabla f_{(x_t, y_t)}(h_t), \sigma^2\big)}[\tilde{\nabla}_t], h_t - h^* \big\rangle \bigg) \cr
    &= \sum_{t=1}^T\big( \E_{\tilde{\nabla}_t\sim\calN\big(\nabla f_{(x_t, y_t)}(h_t), \sigma^2\big)}[f^\prime_t(h_t) - f^\prime_t(h^*)]\big) \cr
    &= \E_{\tilde{\nabla}_t\sim\calN\big(\nabla f_{(x_t, y_t)}(h_t), \sigma^2\big)}\big[\sum_{t=1}^T\big( f^\prime_t(h_t) - f^\prime_t(h^*)\big)\big] \cr
    &= \textit{regret}_T(f^\prime) \cr
    &\le \frac{3}{2}D\sqrt{(L^2+d\sigma^2)T}
\end{align}

Plugin \cref{eq:regret_f_prime} into \cref{eq:regret_f_general} and get:

\begin{equation}
\label{eq:regret_f}
    \err(\bar{h}) \le \err(h^*) + \frac{3D\sqrt{(L^2+d\sigma^2)}}{2\sqrt{T}} + \sqrt{\frac{8 \ln(\nicefrac{2}{\beta^\prime})}{T}} 
\end{equation}

Supposing $\err(h^*) \le \nicefrac{\alpha^\prime}{2}$, in order to get that $\err(\bar{h}) \le \alpha^\prime$ we require that:

\begin{align}
\label{eq:sgd_t}
&\frac{3D\sqrt{(L^2+d\sigma^2)}}{2\sqrt{T}} \le \nicefrac{\alpha^\prime}{4} \Rightarrow T \ge \frac{36(L^2+\sigma^2\sqrt{d})}{\alpha^{\prime2}} \cr
&\text{and} \cr
&\sqrt{\frac{8 \ln(\nicefrac{2}{\beta^\prime})}{T}} \le \nicefrac{\alpha^\prime}{4} \Rightarrow T \ge \frac{32 \ln(\nicefrac{2}{\beta^\prime})}{\alpha^{\prime2}} \cr 
&\text{Therefore:} \cr
&T \ge \max\bigg(\frac{36(L^2+\sigma^2\sqrt{d})}{\alpha^{\prime2}}, \frac{32 \ln(\nicefrac{2}{\beta^\prime})}{\alpha^{\prime2}}\bigg)
\end{align}

\begin{lemma}
\label{lem:sgd-alg-strongly-convex}
Given the settings in \cref{lem:sgd-alg} with the following changes: let $f_{(x, y)}(h) = \Loss(h(x), y)+\frac{\gamma}{2}\|h\|^2$ $L$-lipschitz and $\gamma$-strongly convex loss function and the $\gamma$-convex function $f_t^\prime(h) = \langle \tilde{\nabla}_t, h\rangle + \frac{\gamma}{2}\|h\|^2$, then, with parameter $\eta_t=\frac{D}{\gamma t}$ w.p. at least $\nicefrac{1}{\beta^\prime}$, after $T \ge \tilde{O}\bigg(\max\bigg(\frac{4(L^2+d\sigma^2)}{2\gamma\alpha^\prime}, \frac{32 \ln(\nicefrac{2}{\beta^\prime})}{\alpha^{\prime2}}\bigg)\bigg)$ iterations of \cref{alg:online_sgd}, $\err(\bar{h}) \le \alpha^\prime$.
\end{lemma}

Proof:

The proof is identical to the proof of \cref{lem:sgd-alg} with the following changes: 

According to theorem 3.1 in \cite{Hazan16} $\textit{regret}_T(f^\prime) = \sum_{t=1}^T \big(f^\prime_t(h_t) - f^\prime_t(h^*)\big) \le \frac{(L^2+d\sigma^2)(1+log(T))}{2\gamma}$. Therefore, 

\begin{align}
\label{eq:regret_f_prime_strongly}
    &\textit{regret}_T(f) = \sum_{t=1}^T \big(f_{(x_t, y_t)}(h_t) - f_{(x_t, y_t)}(h^*)\big) \cr
    &\le 
    \sum_{t=1}^T\bigg(\big\langle \nabla f_{(x_t, y_t)}(h_t), h_t - h^* \big\rangle - \gamma\| h_T -h^*\|^2\bigg) \tag{$\gamma$ strongly convexity of $f$} \\
    &\le 
    \sum_{t=1}^T\bigg(\big\langle \nabla f_{(x_t, y_t)}(h_t), h_t - h^* \big\rangle + \gamma\| h_T -h^*\|^2\bigg) \cr
    &\le 
    \sum_{t=1}^T\bigg(\big\langle \nabla f_{(x_t, y_t)}(h_t), h_t - h^* \big\rangle + \gamma\| h_T \|^2 -\gamma\|h^*\|^2\bigg) \cr
    &= \sum_{t=1}^T\bigg(\big\langle \E_{\tilde{\nabla}_t\sim\calN\big(\nabla f_{(x_t, y_t)}(h_t), \sigma^2\big)}[\tilde{\nabla}_t], h_t - h^* \big\rangle \cr 
    &+ \gamma\| h_T \|^2 -\gamma\|h^*\|^2\bigg) \bigg) \cr
    &= \sum_{t=1}^T\big( \E_{\tilde{\nabla}_t\sim\calN\big(\nabla f_{(x_t, y_t)}(h_t), \sigma^2\big)}[f^\prime_t(h_t) - f^\prime_t(h^*)]\big) \cr
    &= \E_{\tilde{\nabla}_t\sim\calN\big(\nabla f_{(x_t, y_t)}(h_t), \sigma^2\big)}\big[\sum_{t=1}^T\big( f^\prime_t(h_t) - f^\prime_t(h^*)\big)\big] \cr
    &= \textit{regret}_T(f^\prime) \cr
    &\le \frac{(L^2+d\sigma^2)(1+log(T))}{2\gamma}
\end{align}

Plugin \cref{eq:regret_f_prime_strongly} into \cref{eq:regret_f_general} and get:

\begin{equation}
\label{eq:regret_f}
    \err(\bar{h}) \le \err(h^*) + \frac{(L^2+d\sigma^2)(1+log(T))}{2\gamma T} + \sqrt{\frac{8 \ln(\nicefrac{2}{\beta^\prime})}{T}} 
\end{equation}

Supposing $\err(h^*) \le \nicefrac{\alpha^\prime}{2}$, in order to get that $\err(\bar{h}) \le \alpha^\prime$ we require that:

\begin{align}
\label{eq:sgd_t}
&\frac{(L^2+d\sigma^2)(1+log(T))}{2\gamma} \le \nicefrac{\alpha^\prime}{4} \cr 
&\Rightarrow \frac{T}{1+log(T)} \ge \frac{4(L^2+d\sigma^2)}{2\gamma\alpha^\prime} \cr
&\Rightarrow T \ge \tilde{O}\bigg(\frac{4(L^2+d\sigma^2)}{2\gamma\alpha^\prime}\bigg) \cr
&\text{and} \cr
&\sqrt{\frac{8 \ln(\nicefrac{2}{\beta^\prime})}{T}} \le \nicefrac{\alpha^\prime}{4} \Rightarrow T \ge \frac{32 \ln(\nicefrac{2}{\beta^\prime})}{\alpha^{\prime2}} \cr 
&\text{Therefore:} \cr
&T \ge \tilde{O}\bigg(\max\bigg(\frac{4(L^2+d\sigma^2)}{2\gamma\alpha^\prime}, \frac{32 \ln(\nicefrac{2}{\beta^\prime})}{\alpha^{\prime2}}\bigg)\bigg)
\end{align}

\begin{lemma}
\label{lem:tcdp_no_sampling}
The Gaussian mechanism $M_G$ in \cref{alg:online_sgd} satisfies $\big(\frac{2L^2}{\sigma^2}, \infty \big)$-tCDP, where $L$ is the lipschitz constant of the function $\Loss(h(x), y)$.
\end{lemma}

Proof:

According to the Lemma 2.4 in \citep{BunS16}, a Gaussian mechanism which answers a sensitivity-$\Delta$ query by adding noise drawn from $\calN(0, \sigma^2)$, satisfies $\big(\frac{\Delta^2}{2\sigma^2}\big)$-zCDP and $\big(\frac{\Delta^2}{2\sigma^2}, \infty \big)$-tCDP. Therefore
The $M_G$ mechanism that runs the query $\nabla f_{(x, y)}(h) = \nabla \Loss(h(x), y)$, and since $\| \nabla \Loss(h(x), y) \| \le L$ where $L$ is the lipschitz constant of $\Loss(h(x), y)$, $M_G$ satisfies $\big(\frac{(2L)^2}{2\sigma^2}, \infty \big)$-tCDP

\begin{lemma}
\label{lem:tcdp_sampling}
Given: a dataset $A$ of size $n$, its neighbor dataset $B$ of size $n-1$, and two sets of weights - $w_A$ and $w_B$ - on theses datasets, where $\forall i=1,...,n-1, w_A[i] = w_B[i]$ and $\sum_{i=1}^{n} w_A[i] \ge \kappa n$. Then, running an one iteration of \cref{alg:online_sgd}, that samples from the distribution that induced by the set of
weights and assigns a Gaussian mechanism, guarantees $\big(\frac{26L^2}{\kappa^2n^2\sigma^2}, \frac{\sigma^2}{8 L^2}\log(\kappa n) \big)$-tCDP, where $\sigma \ge \sqrt{20}L$, and $n \ge O\big(\frac{e}{\kappa}\big)$.
\end{lemma}

The proof of \cref{lem:tcdp_sampling} will given after the following lemma:

\begin{lemma}[Privacy amplification by subsampling - Theorem 12 in \citep{BunDRS18}]
\label{lem:subsampling_tcdp_Bun}

Let $M: \calX^N \to \mathcal{Y}$ satisfy $\rho-zCDP$. Fix $x, x^\prime \in \calX^N$ differing in a single index $i\in [n]$. Let $S \subset [N]$ with $|S| = n$ be uniformly random and let $x_S, x^\prime_S \in \calX^n$ denote the restrictions of $x$ and $x^\prime$, respectively, to the indices in $S$. We define the following probability densities:

\begin{align*}
    P &= \text{ the density of } M(x_S) \text{ conditioned on } i\in S, \\
    Q &= \text{ the density of } M(x^\prime_S) \text{ conditioned on } i\in S, \\
    R &= \text{ the density of } M(x_S) \text{ conditioned on } i \not\in S, \\    
    &= \text{ the density of } M(x^\prime_S) \text{ conditioned on } i \not\in S, \\
\end{align*}

Since index $i$ appears in the sample $S$ with probability $s$, we may write the density of $M(x_S) = sP + (1-s)R$ and $M(x_S) = sP + (1-s)R$.

Given the following conditions hold: 

\begin{enumerate}
\item Let $P,Q,R$ be probability density functions over $\mathcal{Y}$, such that \begin{equation*}
    D_\alpha(P_1\|P_2) \le \rho \alpha
\end{equation*}
for every pair $P_1, P_2 \in \{P,Q,R\}$ and all $\alpha \in (1, \omega^\prime)$.
\item Let $\rho \in (0, 0.1]$.
\item $s\in(0,1]$. $log(\nicefrac{1}{s}) \ge 3\rho(2+\log_2(\nicefrac{1}{\rho}))$.
\item $\omega^\prime = \frac{1}{2\rho}\log(\nicefrac{1}{s}) \ge 3$. 
\end{enumerate}

Then 
\begin{equation*}
    D_\alpha(sP+(1-s)R\|sQ+(1-s)R) \le 13s^2\rho\alpha
\end{equation*}
for every $\alpha\in(1, \nicefrac{\omega^\prime}{2})$.
\end{lemma}

Proof of \cref{lem:tcdp_sampling}:

In each iteration of \cref{alg:online_sgd} Our SGD algorithm does not take a sample in sequence but samples one sample at random from the distribution induced by the weights of the dataset, and then uses the Gaussian mechanism by adding Gaussian noise. We use \cref{lem:subsampling_tcdp_Bun} for enhancing the privacy of the Gaussian mechanism by the subsampling. In the following we see how the all conditions of \cref{lem:subsampling_tcdp_Bun} holds in our case.

Let $\mu_A$ and $\mu_B$ two distributions that induced by the sets of weights $w_A$ and $w_B$ on the datasets. Let $P_A$ and $P_B$ distribution on the output of $M_G\big(f_{(x, y)}(h) \sim \mu_A\big)$ and $M_G\big(f_{(x, y)}(h) \sim \mu_B\big)$ respectively, depending on the randomness of the $M_G$ and the sampling from the distributions. 
We define the following probability densities of the output of the mechanism $M_G$ conditioned on the sampling of the single example:

\begin{align*}
    P &= \text{ the density of } M_G\big(f_{(x_n, y_n)}(h)\big) \text{ from } \mu_A,\\
    Q &= \text{ the density of } M_G\big(f_{(x_n, y_n)}(h)\big) \text{ from } \mu_B, \\
    R &= \text{ the density of } M_G\big(f_{(x_i, y_i)}(h) \sim \mu_A | i \ne n\big)\\
    &= \text{ the density of } M_G\big(f_{(x_i, y_i)}(h) \sim \mu_B\big).
\end{align*}

The density $Q$ cannot be really occur in our case as the probability to get $(x_n, y_n)$ from $\mu_B$ is 0, but for formalization of the proof we assume that $Q = R$.

We get that, 
\begin{align*}
&D_\alpha(P\|Q) = 1 \\
\text{and} \\
&D_\alpha(Q\|R) = D_\alpha(P\|R) \\
&= D_\alpha\bigg(M_G\big(f_{(x_n, y_n)}(h)\big) \| M_G\big(f_{(x_i, y_i)}(h) \sim \mu_A | i \ne n\big)\bigg) \\
&= D_\alpha\bigg(M_G\big(f_{(x_n, y_n)}(h)\big) \bigg\| \E_{(x_i, y_i) \sim  \mu_A | i \ne n}\bigg[M_G\big(f_{(x_i, y_i)}(h)\big)\bigg]\bigg) \\
\end{align*}

from quasi-convex property of zCDP \cref{TO COMPLETE} we get 
\begin{align*}
    &\le \max_i D_\alpha\bigg(M_G\big(f_{(x_n, y_n)}(h)\big) \bigg\| M_G\big(f_{(x_i, y_i)}(h)\big)\bigg) \\
    & \underset{\substack{\text{\cref{lem:tcdp_no_sampling}}}}{\le}
    (\frac{2L^2}{\sigma^2}\cdot\alpha)\text{-ZCDP} \\
    &= (\frac{2L^2}{\sigma^2}\cdot\alpha, \infty)\text{-ZCDP}
\end{align*}

Therefore \textbf{condition 1} holds with $\rho= \frac{2L^2}{\sigma^2}$ and $\omega^\prime = \infty$.

We get $\rho = \frac{2L^2}{\sigma^2}$, Therefore \textbf{Condition 2} $\rho \in (0, 0.1]$ holds when $\frac{2L^2}{\sigma^2} \le 0.1 \Rightarrow \sigma \ge \sqrt{20}L$.

\textbf{Condition 3:} According to our settings that $\sum_{i=1}^n \omega_A[i] \ge \kappa n$ the probability $s$ to sample the $(x_n, y_n)$ is $s \le \frac{1}{\kappa n}$. Therefore condition 3) $\log_2(\nicefrac{1}{s}) \ge 3\rho(2+\log_2(\nicefrac{1}{\rho}))$ in the worst case where 
$s = \frac{1}{\kappa n}$ holds when 

\begin{align*}
&\log_2(\kappa n) \ge 3\rho(2+\log_2(\nicefrac{1}{\rho})) \\
&\Rightarrow n \ge \frac{e^{6\rho}\cdot  e^{3\rho\log_2(\nicefrac{1}{\rho})}}{\kappa} \\
&= \Omega\bigg( \frac{e^{6\rho}\cdot(\frac{1}{\rho})^{3\rho}}{\kappa} := f(\rho)\bigg)
\end{align*}

For $f(1)$ we get $n = \frac{e^6}{\kappa}$ and it is correct also for smaller values $0\le \rho \le 1$, because $f^\prime(\rho) = \frac{(1-\ln(\rho))3e^{6\rho}}{\rho^{3\rho}}$ is positive in $(0, 1]$. Therefore also for $\rho \le 0.1$ the requirement holds with $n = \Omega(\frac{1}{\kappa})$.

The \textbf{condition 4} $\omega^\prime = \frac{1}{2\rho}\log(\nicefrac{1}{s}) \ge 3$ can achieved by truncation of our $w^\prime = \infty$ to be $w^\prime = \frac{1}{2\rho}\log(\nicefrac{1}{s}) \ge 3$. According our settings that $s \le \nicefrac{1}{\kappa n}$, when $n \ge \frac{e^{6\rho}}{\kappa}$
we get $ \frac{1}{2\rho}\log(\nicefrac{1}{s}) \ge \frac{1}{2\rho}\log(\kappa n) \ge 3$ as required.

Therefore $n \ge \frac{e^{6\rho}}{\kappa} = \frac{e^{\nicefrac{12 L^2}{\sigma^2}}}{\kappa}$.

Given all the requirements hold as before, according to \cref{lem:subsampling_tcdp_Bun}, for all $\alpha \le \nicefrac{\omega^\prime}{2} = \frac{1}{4\rho}\log(\nicefrac{1}{s})$ (therefore at least for $\alpha \le \frac{1}{4\rho}\log(\kappa n) $:

\begin{align*}
    &D_\alpha\bigg(M_G\big(f_{(x, y)}(h) \sim \mu_A\big) \| M_G\big(f_{(x, y)}(h) \sim \mu_B\big) \bigg) \\
    &= D_\alpha\bigg(sP + (1-s)R  \| R \bigg)\\
    &= D_\alpha\bigg(sP + (1-s)R  \| sR + (1-s)R \bigg)\\
    & \underset{\substack{\text{Q=R}}}{=}
    D_\alpha\bigg(sP + (1-s)R  \| sQ + (1-s)R \bigg)\\
    &\le 13s^2\rho \alpha \le \frac{26 L^2  \alpha}{\sigma^2 \kappa^2 n^2}
\end{align*}

Where the last inequality comes by the settings $\rho = \frac{2L^2}{\sigma^2}$ (\cref{lem:tcdp_no_sampling}) and $s \le \frac{1}{\kappa n}$.

\begin{lemma}
\label{lem:online_sgd_composition}
For convex loss function, after running $O\bigg(\frac{\log(\nicefrac{1}{\kappa})}{\alpha^2}\bigg)$ times the \cref{alg:online_sgd}  with $\max\bigg(\frac{36(L^2+\sigma^2\sqrt{d})}{\alpha^2}, \frac{32 \ln(\nicefrac{2}{\beta})}{\alpha^2}\bigg)$ iterations in each run, then, according composition theorem of tCDP \cref{To complete}, composition of $\big(\frac{26L^2}{\kappa^2n^2\sigma^2}, \frac{\sigma^2}{4 L^2}\log(\kappa n) \big)$-tCDP, $O\bigg(\frac{\log(\nicefrac{1}{\kappa})}{\alpha^4} \cdot \max\big(36(L^2+\sigma^2\sqrt{d}), 32 \ln(\nicefrac{2}{\beta})\big) \bigg)$ times, guarantees to $(\rho, \omega)$-tCDP with the parameters:

\begin{align}
&\rho = O\bigg(\frac{26L^2}{\kappa^2n^2\sigma^2} \cdot \frac{\log(\nicefrac{1}{\kappa})}{\alpha^4} \cdot \max\big(36(L^2+\sigma^2\sqrt{d}), 32 \ln(\nicefrac{2}{\beta})\big)\bigg)\\ 
&\omega = \frac{\sigma^2}{4 L^2}\log(\kappa n) 
\end{align}

Therefore,  
\begin{equation}
n \ge O\bigg(\sqrt{\frac{26L^2}{\rho\kappa^2n^2\sigma^2} \cdot \frac{\log(\nicefrac{1}{\kappa})}{\alpha^4} \cdot \max\big(36(L^2+\sigma^2\sqrt{d}), 32 \ln(\nicefrac{2}{\beta})\big)}\bigg)
\end{equation}

In the similar way, for $\gamma$-strongly convex function, we get: 

\begin{align}
\rho = \tilde{O}\bigg(\frac{26L^2}{\kappa^2n^2\sigma^2} \cdot \frac{\log(\nicefrac{1}{\kappa})}{\alpha^2} \cdot \max\bigg(\frac{4(L^2+d\sigma^2)}{2\gamma\alpha}, \frac{32 \ln(\nicefrac{2}{\beta})}{\alpha^2}\bigg)\bigg)
\end{align}

According to lemma 6 in \cite{BunDRS18}, for all $\delta>0$ and all $1< \alpha \le \omega$, the algorithm satisfies $(\epsilon, \delta)$-differencial privacy with 
\begin{equation}
    \epsilon = 
    \begin{cases}
        \rho+2\sqrt{\rho \log(\nicefrac{1}{\delta})},& \text{if } \log(\nicefrac{1}{\delta}) \le (\omega -1)^2\rho\\
        \rho\omega+\frac{log(\nicefrac{1}{\delta})}{\omega-1},              & \text{otherwise}
    \end{cases}
\end{equation}
\end{lemma}

\section{Specific $\chi^2$-Divergences}
\label{sec:example_chi_square_divergence}

\paragraph{$k$-out of $d$ Larger-Variance Gaussians.}
Consider the specific case where $\S = {\cal N}(0,1)$ and $\T = {\cal N}(0,\sigma^2)$. We thus have that 
\begin{align*}
    \chisq{\T}{\S}+1 &= \int_{x\in\R} \frac{\pdfT(x)^2}{\pdfS(x)}dx 
    \cr &= \frac{\sqrt{2\pi}}{2\pi\sigma^2}\int_{x\in\R} \exp( -\frac {x^2}{2} \left( \frac 2 {\sigma^2} - 1 \right) )dx
    \cr &= \frac {1}{\sigma^2\sqrt{2\pi}} \int_{x\in\R}\exp(-\frac {x^2}{2\cdot \left( \frac{\sigma^2}{2-\sigma^2}  \right)})dx
    \cr&= \frac{\sqrt{2\pi \left( \frac{\sigma^2}{2-\sigma^2}  \right) }}{\sigma^2\sqrt{2\pi}} = \sqrt{\frac 1 {\sigma^2(2-\sigma^2)}}
\end{align*}
Setting $\sigma^2 = 1+\frac{4}{5}=1.8$, we get $\chisq{{\cal N}(0,\sigma^2)}{{\cal N}(0,1)} = \sqrt{\frac{1}{(1+\frac{4}{5})(1-\frac{4}{5})} } = \frac{5}3$.

Now consider the $d$-dimensional case where $\S = {\cal N}(0,I_d)$ and $\T$ is a product distribution where on $k$ coordinates we sample a value from ${\cal N}(0,1.8)$ on the remaining $d-k$ coordinates we sample a value from ${\cal N}(0,1)$. It follows that
\[ \chisq{\T}{\S}+1 = (\nicefrac 5 3)^k \] 

\paragraph{$t$-Equal Features.} Consider $\S$ the distribution which results from the following randomized algorithm, which takes as input a set $P$ of $t$ disjoint pairs of coordinates. (Thus, $t\leq \nicefrac d 2$) and a parameter $\eta\in(0,\nicefrac 1 2)$: (1) Sample each of the $d-2t$ coordinates not in $P$ u.a.r, (2) for each pair $(i,j)\in P$ sample $x_i$ u.a.r and set $x_j=x_i$ w.p. $\tfrac 1 2 + \eta$ and $x_j=1-x_i$ w.p. $\tfrac 1 2 -\eta$. 
Setting $\T$ as the uniform distribution over $\{0,1\}^d$, it is clear that for any $x$ where there are precisely $k$ pairs of coordinates in $P$ where $x_i\neq x_j$, it holds that $w(x) = \frac {2^{-d}}{2^{-(d-t)}(\tfrac 1 2-\eta)^k (\tfrac 1 2 + \eta)^{t-k}} = \left(\frac{1}{2(\tfrac 1 2 + \eta)}\right)^t \left(\frac{\tfrac 1 2 + \eta}{\tfrac 1 2 -\eta}\right)^k$ . It follows that 
\begin{align*}
    \chisq{\T}{\S}+1&= \E_{x\sim \T}[w(x)] 
    \cr &= \sum_{k=0}^t \sum_{x \textrm{ has $k$ disagreeing pairs in $P$}} 2^{-d} \left(\frac{1}{1+2\eta}\right)^t \left(\frac{\tfrac 1 2 +\eta}{\tfrac 1 2 -\eta }\right)^k
    \cr & = \sum_{k=0}^t 2^{-d}2^{d-t}\binom t k\left(\frac{1}{1+2\eta}\right)^t \left(\frac{1+2\eta}{1-2\eta}\right)^k 
    \cr &= \frac{2^{-t}}{(1+2\eta)^t} \sum_{k=0}^t\binom t k \left(\frac{1+2\eta}{1-2\eta}\right)^k 
    \cr &= \frac{2^{-t}}{(1+2\eta)^t} (1+\frac{1+2\eta}{1-2\eta})^t = 
    \cr &= \frac{2^{-t}}{(1+2\eta)^t} \cdot \frac{2^{t}}{(1-2\eta)^t} = (\frac {1}{1-4\eta^2})^t 
\end{align*}
In contrast, $\Dinfty{\T}{\S} = \left(\frac{1}{1+2\eta}\right)^t \left(\frac{\tfrac 1 2 +\eta}{\tfrac 1 2 -\eta }\right)^t = (\frac{1}{1-2\eta})^t$. 
Setting $\eta = \nicefrac {t-1}{2t}$ we get $\chisq{\T}{\S} = (\frac {t^2} {t^2-{(t-1)^2}})^t = (\tfrac {2t+1} 4 + \tfrac{1}{8t-4})^t\approx (\tfrac t 2)^t$. In contrast, $\Dinfty{\T}{\S} = t^t$.

}

\end{document}